\documentclass[runningheads]{llncs}
\usepackage{graphicx}
\usepackage{mwe} 
\usepackage{capt-of}
\usepackage{float}
\usepackage{amsmath}
\usepackage[misc,geometry]{ifsym}

\usepackage[inline]{enumitem}
\usepackage{comment}
\usepackage{multirow}
\usepackage{amssymb}
\usepackage{subcaption}
\usepackage{xcolor}

\makeatletter
        
        \def\env@closedcases{%
            \let\@ifnextchar\new@ifnextchar
            \left\{
            \def\arraystretch{1.2}%
            \array{@{}l@{\quad}l@{}}%
        }
\makeatother

    \makeatletter
\def\@fnsymbol#1{\ensuremath{\ifcase#1\or \dagger\or \ddagger\or
   \mathsection\or \mathparagraph\or \|\or **\or \dagger\dagger
   \or \ddagger\ddagger \else\@ctrerr\fi}}
    \makeatother

\begin{document}
\title{ Histopathological Image Analysis with Style-Augmented Feature Domain Mixing for Improved Generalization}
\titlerunning{FuseStyle}
%
\author{Vaibhav Khamankar\thanks{These authors contributed equally to this work and share the first authorship} \and Sutanu Bera$^\dagger$ \Letter  \and Saumik Bhattacharya \and Debashis Sen \and Prabir Kumar Biswas}


\authorrunning{V.Khamankar et al.}
%
\institute{Department of Electronics and Electrical Communication Engineering \\ Indian Institute of Technology Kharagpur,  India.\\
\email{sutanu.bera@iitkgp.ac.in}}
\maketitle              
\begin{abstract}
Histopathological images are essential for medical diagnosis and treatment planning, but interpreting them accurately using machine learning can be challenging due to variations in tissue preparation, staining and imaging protocols. Domain generalization aims to address such limitations by enabling the learning models to generalize to new datasets or populations. Style transfer-based data augmentation is an emerging technique that can be used to improve the generalizability of machine learning models for histopathological images. However, existing style transfer-based methods can be computationally expensive, and they rely on artistic styles, which may negatively impact model accuracy. In this study, we propose a feature domain style mixing technique that uses adaptive instance normalization to estimate style-mixed versions of image features. We compare our proposed method with existing style transfer-based data augmentation methods and found that it performs similarly or better, despite requiring lower computation. Our results demonstrate the potential of feature domain statistics mixing in the generalization of learning models for histopathological image analysis.


\end{abstract}
\section{Introduction}

Histopathological images play a critical role in medical diagnosis and treatment planning, allowing healthcare providers to visualize the microscopic structures of tissues and organs. However, accurately interpreting these images can be challenging due to variations in tissue preparation, staining and imaging protocols. These variations can result in significant differences in image quality, tissue morphology and staining intensity, making it difficult to develop machine learning models for analysis that generalize well to new datasets or populations. Domain generalization is a field of machine learning that seeks to address this limitation by enabling models to generalize to new domains or datasets. In the context of histopathological images, domain generalization methods aim to improve the generalizability of machine learning models by reducing the effects of dataset bias and increasing the robustness of the model to variations in tissue preparation, staining, and imaging protocols. Recently, there has been a growing interest in using style transfer-based data augmentation for learning visual representations that are independent of specific domains for histopathological images viz., \cite{staingan,stainnet,STRAP}. This technique involves transferring the style or texture of one image to another while maintaining the original content. By generating new images with different styles or textures, this technique can be used to augment the training data and improve the model's generalization performance \cite{STRAP}. 
Although the style transfer based method achieves good results in domain generalization for histopathological images, it takes a considerable amount of time to generate the augmented data. Further, the collinearity between the various artistic styles used for the style transfer may have a negative impact on the model's accuracy. 

Unlike the existing methods, in this work, we propose to apply feature domain style mixing for the style transfer. Specifically, we use adaptive instance normalization \cite{adain} to mix the feature statistics of the different images to generate a style-augmented version of an image. Feature statistics mixing helps to save a lot of time and computation power as data augmentation is not required, and the dependency on the artistic style is also alleviated. We compare the proposed method with the current state-of-the-art style transfer-based data augmentation methods, on two image classification tasks and one object detection task. We find that the proposed method performs similarly or better than the image domain mixing-based methods, despite having low computation requirements.

\section{Related Work}
In the field of digital pathology, researchers have developed several deep learning approaches to address challenges related to domain generalization such as normalization and style transfer. One example is StainNet \cite{stainnet}, which is designed for stain normalization in digital pathology images. StainNet removes variations in tissue staining across different samples, making it easier to compare and analyze images in a consistent manner. Another approach, STRAP \cite{STRAP}, uses a deep neural network to extract features from histopathology images and proposes a style transfer augmentation technique to reduce the domain-specific information in these features. This technique generates a new set of images that have the same content as the original images but in different styles. Domain Adversarial RetinaNet \cite{DAR}, a modified version of the RetinaNet object detection model, has been developed that includes domain adversarial training. The idea is to train in both source and target domain data to address domain generalization challenges. 

\section{Proposed Method}
\subsection{Background}
Huang et al. \cite{adain} introduced Adaptive Instance Normalization (AdaIN) for style transfer based on Instance normalization \cite{IN}. AdaIN aims to align the means and variances of instances of the content features ($c$) with those of the style features ($s$). It computes the mean ($\mu(s)$) and variance ($\sigma(s)$) parameters from instances of the style input and achieves the style transfer as $AdaIN(c) = {\sigma(s)\frac{x-\mu(c)}{\sigma(c)}+\mu(s)}$
, where $\mu(c)$ and $\sigma(c)$ are respectively the corresponding instance mean and standard deviation of a given content feature tensor. The above parameter adaption allows for arbitrary style transfer, enabling the mixing of the content and style features in a way that produces a new output with the parameters of the style.

\subsection{FuseStyle: Proposed feature domain style mixing}
Our feature domain style mixing approach, FuseStyle, is inspired by AdaIN. ~FuseStyle avoids the use of an image generating network that is usually associated with style transfer based domain generalization. Instead, it regularizes the training of the neural network at hand (for performing a required task) by perturbing the style information of the training instances. It can be easily implemented as a plug-and-play module inserted between the layers of the neural network. So. the need to explicitly create a new style image does not arise.

\begin{figure}[tbp]
    \centering
    \includegraphics[width=\textwidth]{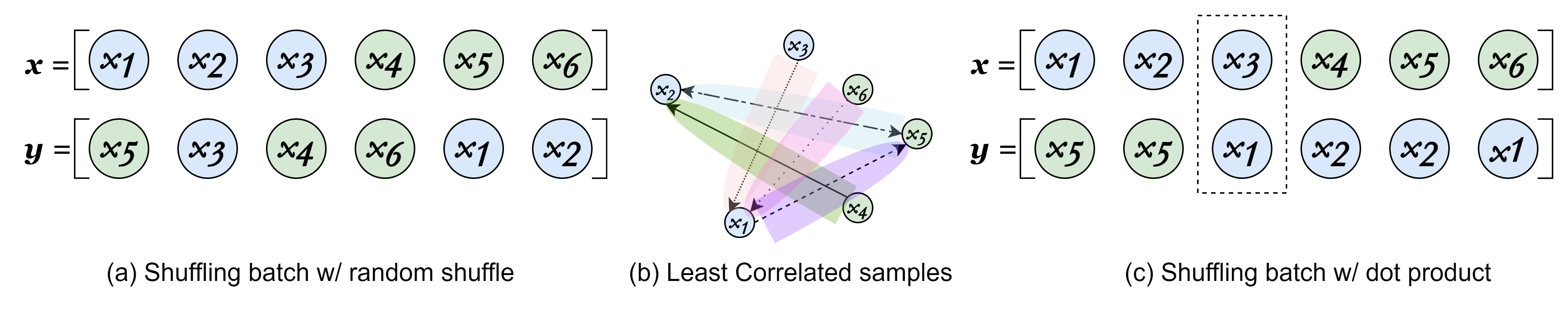}
    \caption{A graphical illustration of FuseStyle. The shaded areas in (b) are the simulated points for augmentation. The domain label of each sample is colour-coded. There can be cases where the dot product (correlation) is the least within the domain as highlighted in the dotted rectangle in (c).}
    \label{fig:net}
\end{figure}

FuseStyle, depicted in Fig.\ref{fig:net}, combines the feature statistics of two instances from the same /different domains as a convex sum using random weights to simulate new styles.
As shown in Fig.\ref{fig:net}, for an input training batch, $x$, a reference batch $y$ is generated by shuffling $x$ across the batch dimension. We then compute the means ($\mu$) and variances ($\sigma$) of the corresponding instances in $x$ and $y$, and use them to compute the combined feature statistics as:
\begin{equation}
\label{eqn2}
\gamma_i = {\lambda_i\sigma(x_i) + (1-\lambda_i)\sigma(y_i)},\;\;\;\;\;\;\beta_i = {\lambda_i\mu(x_i) + (1-\lambda_i)\mu(y_i)}
\end{equation}
where $i$ denotes the $i^{\textrm{th}}$ instance and $\lambda_i\sim\textrm{Beta}(\alpha,\alpha)$ is computed from a Beta distribution having both its shape parameters as $\alpha$. 
A style-modified training instance $\Tilde{x_i}$ is then computed as:	\begin{equation} \label{eqn4}
	\Tilde{x_i} = \gamma_i\frac{x_i-\mu(x_i)}{\sigma(x_i)}+\beta_i
	\end{equation}
where the batch size of $\Tilde{x}$ is the same as that of $x$ and $y$. $x$ is then randomly (binomial-$B(0,.5)$) replaced by $\Tilde{x}$ as the training batch for domain generalization.

 
Generating the reference batch $y$ is crucial for achieving better generalization to unseen domains. While previous studies \cite{mixstyle} have used a random sample selection method for creating the reference batch, a recent study \cite{STRAP} in histopathological image domain generalization has shown that mixing medically irrelevant images, such as artistic paintings, with whole slide images (WSI) results in improved performance. This suggests that using the least correlated image in the reference batch could result in a better generalization than using a meaningful stylized image. With this motivation, we propose a new method of generating the reference batch that allows the mixing of the features of a sample with the features of another sample in the batch that is least correlated to the former. This method has inherent advantages over existing methods. For example, when we combine the parameters of two furthest samples linearly, the interpolated parameter values are more likely to represent a simulated sample that is far from the both the original samples than when we combine two close samples (which may happen during random reference batch generation). This allows us to explore more regions in the feature space and simulate a wider variety of augmented domains, as illustrated in Figure \ref{fig:net}b. Consider that FuseStyle is applied between a layer, $f_{l}$, and $f_{l+1}$, and the output feature of the layer $f_l$ is $z_l \in \mathbb{R}^{B \times C \times W \times H}$ ($B$ - batch dimension). Then, the correlation ($\rho  \in \mathbb{R}^{B \times B}$) between different samples of the current batch can be computed by:
\begin{equation}
    \rho = \hat{z}_l \odot \hat{z}_l^T
\end{equation}
where $\odot$ represents the matrix multiplication, $\hat{z}_l \in \mathbb{R}^{B \times CWH}$ is the vectorized version of the $z_l$ and $T$ represents the transpose operation. Next, we set $i^{th}$ sample of the reference batch, that is, $y_i$ to be $x_j$, where $j = \arg \min_j\rho_i$, and $\rho_i \in \mathbb{R}^B$ is the $i^{th}$ row of the matrix $\rho$. Then, the $i^{th}$ sample of the batch $x$ is mixed with $i^{th}$ sample of the batch $y$ as mentioned in Eq.(\ref{eqn4}) to get $\Tilde{x_i}$. We set $\alpha$ of the Beta distribution to 0.3 to generate all the results reported in this paper. During the learning phase of the neural network model, the probability of using the FuseStyle method is set at 0.5, but it is not applied during the test phase.
\begin{table}[tbp]
\caption{Comaprison of FuseStyle with SoTA methods on Camelyon17-WILDS.}
\resizebox{\textwidth}{!}{
\centering
\begin{tabular}{|c|c|c|c|c|c|c|c|c|c|}
\hline
Methods       & STRAP  & FuseStyle & LISA   & Fish   & ERM    & V-REx  & DomainMix & IB-IRM & GroupDRO \\ \hline
Test Accuracy & 93.7\% & 90.49\%   & 77.1\% & 74.7\% & 70.3\% & 71.5\% & 69.7\%    & 68.9\% & 68.4\%   \\ \hline
\end{tabular}}
\label{tab:CAMSOTA}
\end{table}



\section{Experimental Details}
\subsection{Datasets and Task}
In our study, we compared our proposed method with the recent state-of-the-art histopathological domain generalization using two datasets. 1. The MIDOG'21 Challenge dataset\cite{midog21} consisted of 200 samples of human breast cancer tissue stained with Haematoxylin and Eosin (H\&E). Four scanning systems were used to digitize the samples: Leica GT450, Aperio CS2 (CS), Hamamatsu XR (XR), and Hamamatsu S360 (S360), resulting in 50 WSIs from each system. 2. The Camelyon17-WILDS \cite{wilds} dataset comprised 1,000 histopathology images distributed across six domains, representing different combinations of medical centres and scanners.
In our study, we focused on three tasks: classification between mitotic figures and non-mitotic figures using the MIDOG'21 dataset, tumour classification using the Camelyon17 dataset, and detection of mitotic and non-mitotic figures using the MIDOG'21 dataset. TFor the mitotic figure detection task, the details regarding dataset preparation can be found in the supplementary material of our study. For the Camelyon17 WILDS dataset, we used the default settings and train test split as given on the challenge website. For the classification task on MIDOG'21, we cropped patches of size $ 64 \times 64$ around the mitotic and non-mitotic figure, and we then performed an 80-20 train-test split on the cropped patches keeping the patches from each domain separate. 
\begin{table}[t]
\centering
\caption{Classification using FuseStyle and STRAP on $MIDOG'21$ Dataset.}
\begin{tabular}{|l|lll|lll|}
\hline
                                                                  & \multicolumn{3}{l|}{\begin{tabular}[c]{@{}l@{}}STRAP\\ Test Accuracy(\%)\end{tabular}}     & \multicolumn{3}{l|}{\begin{tabular}[c]{@{}l@{}}FuseStyle\\ Test Accuracy(\%)\end{tabular}} \\ \hline
\begin{tabular}[c]{@{}l@{}}Networks\\ \{Train,Test\}\end{tabular} & \multicolumn{1}{l|}{XR}             & \multicolumn{1}{l|}{S360}           & CS             & \multicolumn{1}{l|}{XR}             & \multicolumn{1}{l|}{S360}           & CS             \\ \hline
\{S360+CS,XR\}                                                    & \multicolumn{1}{l|}{67.33}          & \multicolumn{1}{l|}{87.99}          & 84.67          & \multicolumn{1}{l|}{\textbf{77.56}} & \multicolumn{1}{l|}{\textbf{91.07}} & \textbf{90.46} \\ \hline
\{XR+CS, S360\}                                                   & \multicolumn{1}{l|}{88.35}          & \multicolumn{1}{l|}{\textbf{76.78}} & 91.14          & \multicolumn{1}{l|}{\textbf{90.06}} & \multicolumn{1}{l|}{75.16}          & \textbf{92.16} \\ \hline
\{XR+S360,CS\}                                                    & \multicolumn{1}{l|}{\textbf{88.92}} & \multicolumn{1}{l|}{\textbf{92.70}} & \textbf{74.28} & \multicolumn{1}{l|}{86.65}          & \multicolumn{1}{l|}{88.96}          & 74.10          \\ \hline
\end{tabular}
\label{tab:MDG21}
\end{table}
\subsection{Model Architecture, Training and Methods}
\textbf{Classification:} Here, we employ ResNet50 \cite{resnet} CNN architecture and integrate FuseStyle after layers 1 and 4 of the network for 15 epochs. We use Binary Cross Entropy (BCE) Loss for training, while Adam Optimizer \cite{adam} with a learning rate of 1e-4 is utilized. To facilitate smooth training, a scheduler is used, that is, when no improvement is seen during model training after 2 epochs, the learning rate is reduced by a factor of 0.01. The batch size is set to 256 for both Camelyon17-WILDS \cite{wilds} and MIDOG'21 Challenge datasets \cite{midog21}. Recent studies on style transfer indicate that style information can be modified by altering the instance-level feature statistics in the lower layers of a Convolutional Neural Network (CNN) while preserving the image's semantic content representation~\cite{adain,Dumoulin}, and hence, we consider layers 1 and 4 of the ResNet to use FuseStyle.

\textbf{Mitotic figure Detection:}
For mitotic figure detection, we utilize RetinaNet \cite{retinanet} with ResNet50 as the backbone architecture and incorporate FuseStyle on layers 1 and 4 of the backbone. We use Focal Loss and train the network for 100 epochs on the MIDOG'21 Challenge dataset with a batch size of 6. Adam Optimizer \cite{adam} is used with a learning rate of 1e-4. We use the adaptive learning rate decay scheduler, that is, when no improvement is seen in model training after two epochs, the learning rate is reduced by factor of 0.1 for stable training.

\textbf{Methods:} To assess the effectiveness of our proposed approach, we compare it to eight state-of-the-art domain generalization methods, namely STRAP \cite{STRAP}, LISA \cite{lisa}, Fish \cite{fish}, ERM \cite{erm}, V-REx \cite{vrex}, DomainMix \cite{domainmix}, IB-IRM \cite{ibirm}, and GroupDRO \cite{groupdro} for classification task on the Camelyon17-WILDS dataset, where we evaluate the classification accuracy. The best existing approach STRAP \cite{STRAP} based on the performance data on Camelyon17-WILDS dataset is used further for comparison with the proposed approach on the MIDOG'21 Challenge dataset, where both classification and mitotic figure detection are considered. We implemented the networks using the PyTorch library in Python and utilized a GeForce GTX 2080Ti GPU for efficient processing.
\vspace{-5pt}
\begin{figure}[t]
\centering
\begin{subfigure}{0.2\textwidth}
    \includegraphics[width=0.8\textwidth]{{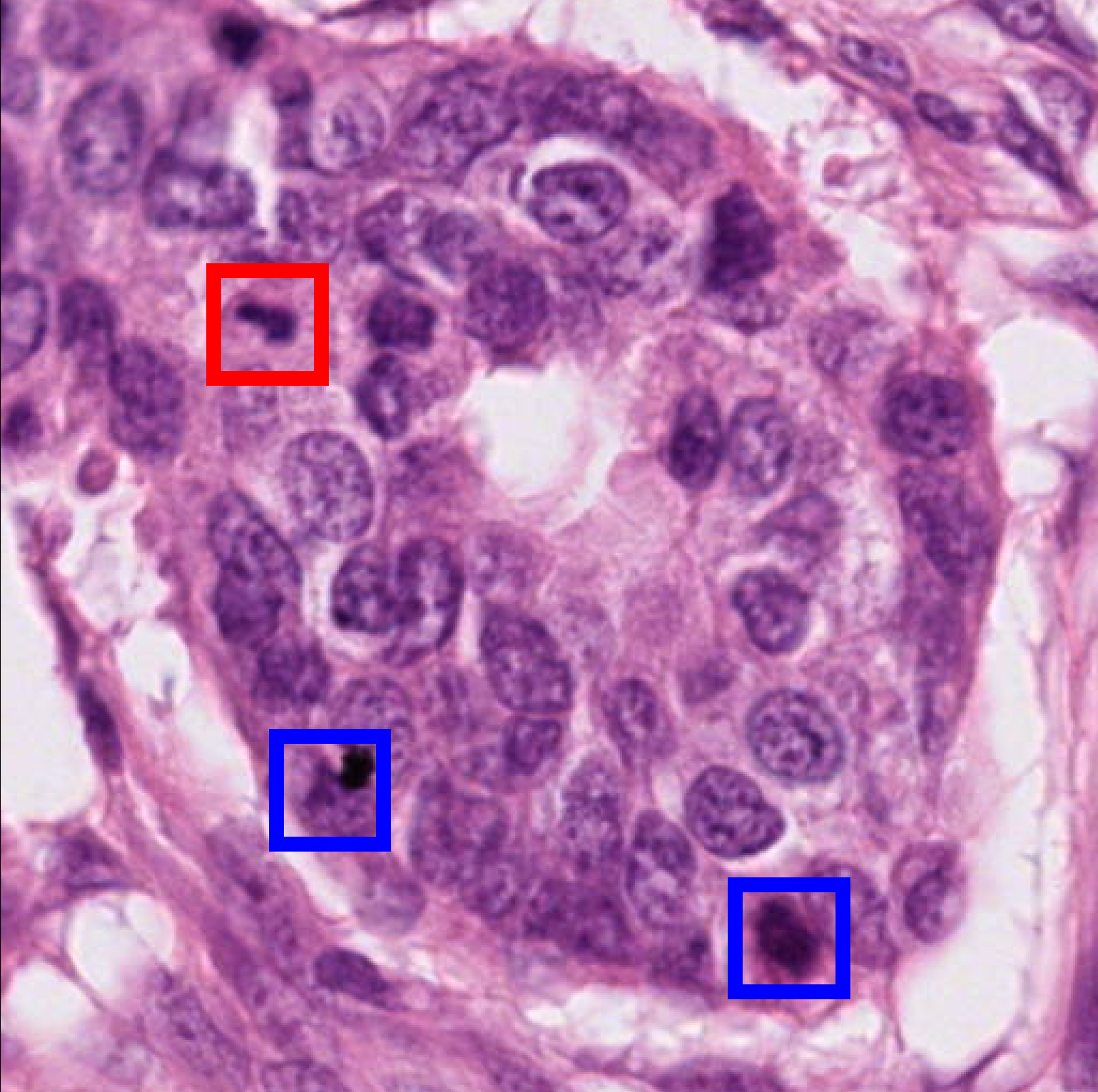}}
    \caption{GT}
    \label{fig:GT}
\end{subfigure}
\hfill
\begin{subfigure}{0.2\textwidth}
    \includegraphics[width=0.8\textwidth]{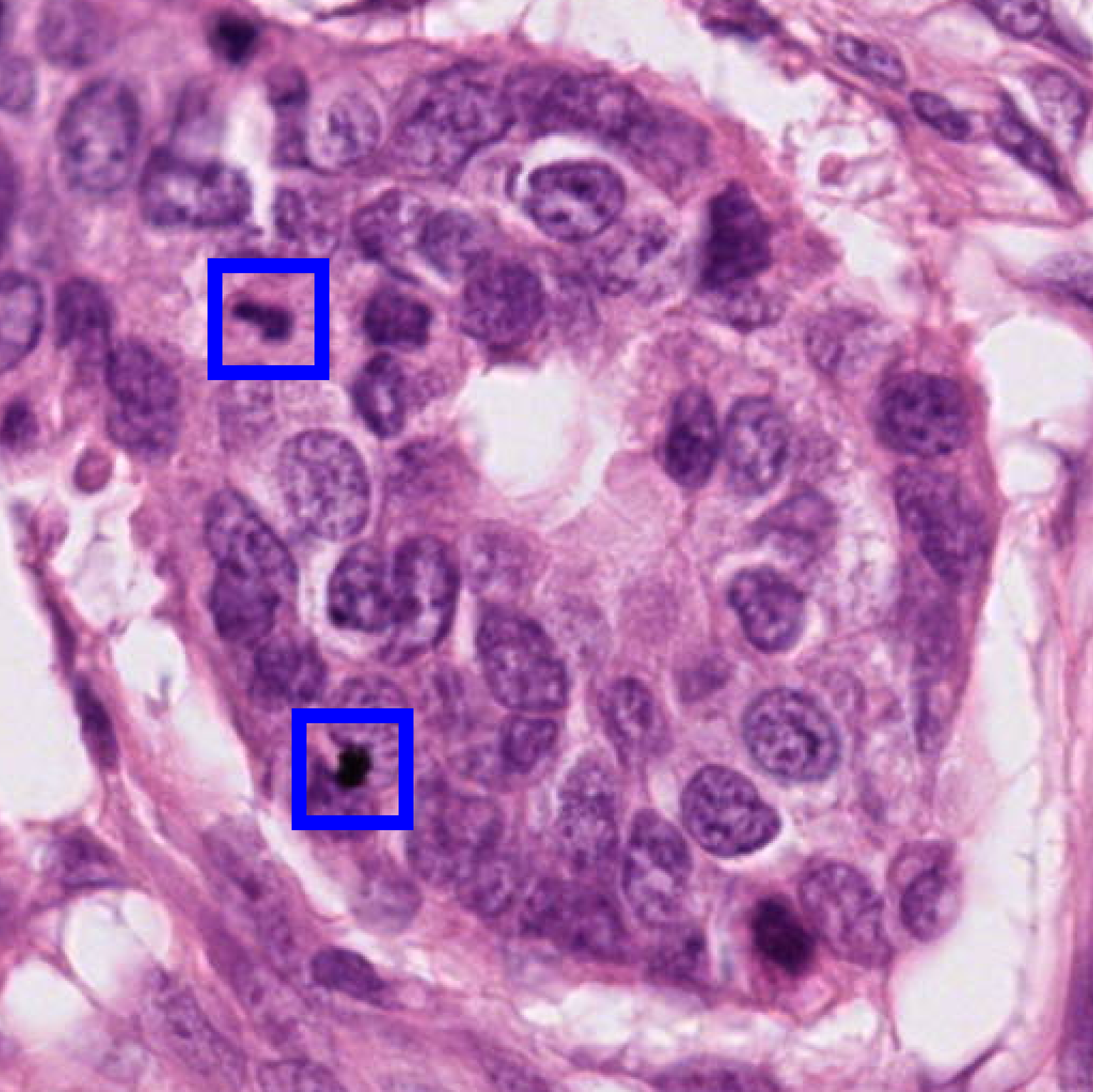}
    \caption{RetinaNet}
    \label{fig:WF}
\end{subfigure}
\hfill
\begin{subfigure}{0.2\textwidth}
    \includegraphics[width=0.8\textwidth]{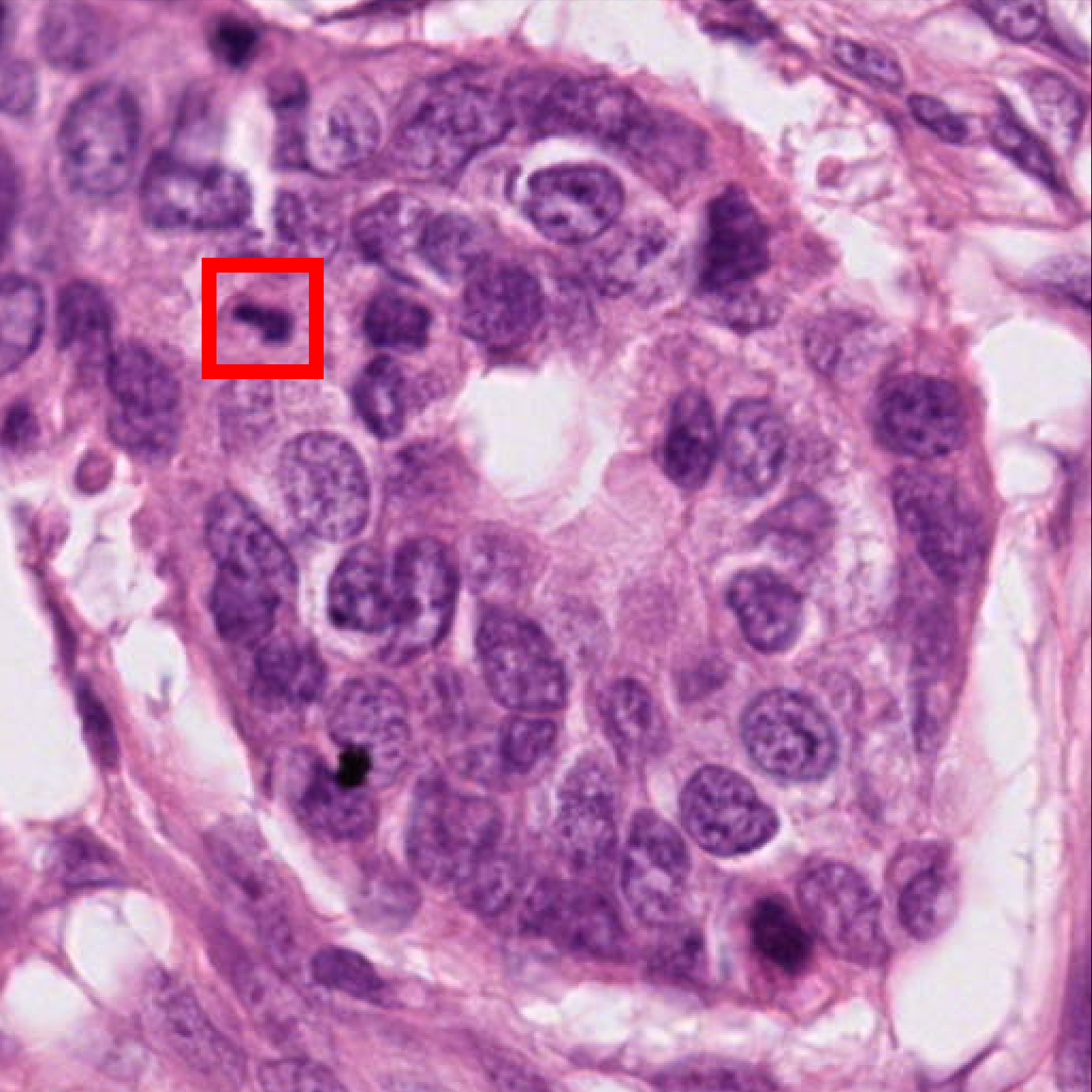}
    \caption{STRAP}
    \label{fig:strap}
\end{subfigure}
\hfill
\begin{subfigure}{0.2\textwidth}
    \includegraphics[width=0.8\textwidth]{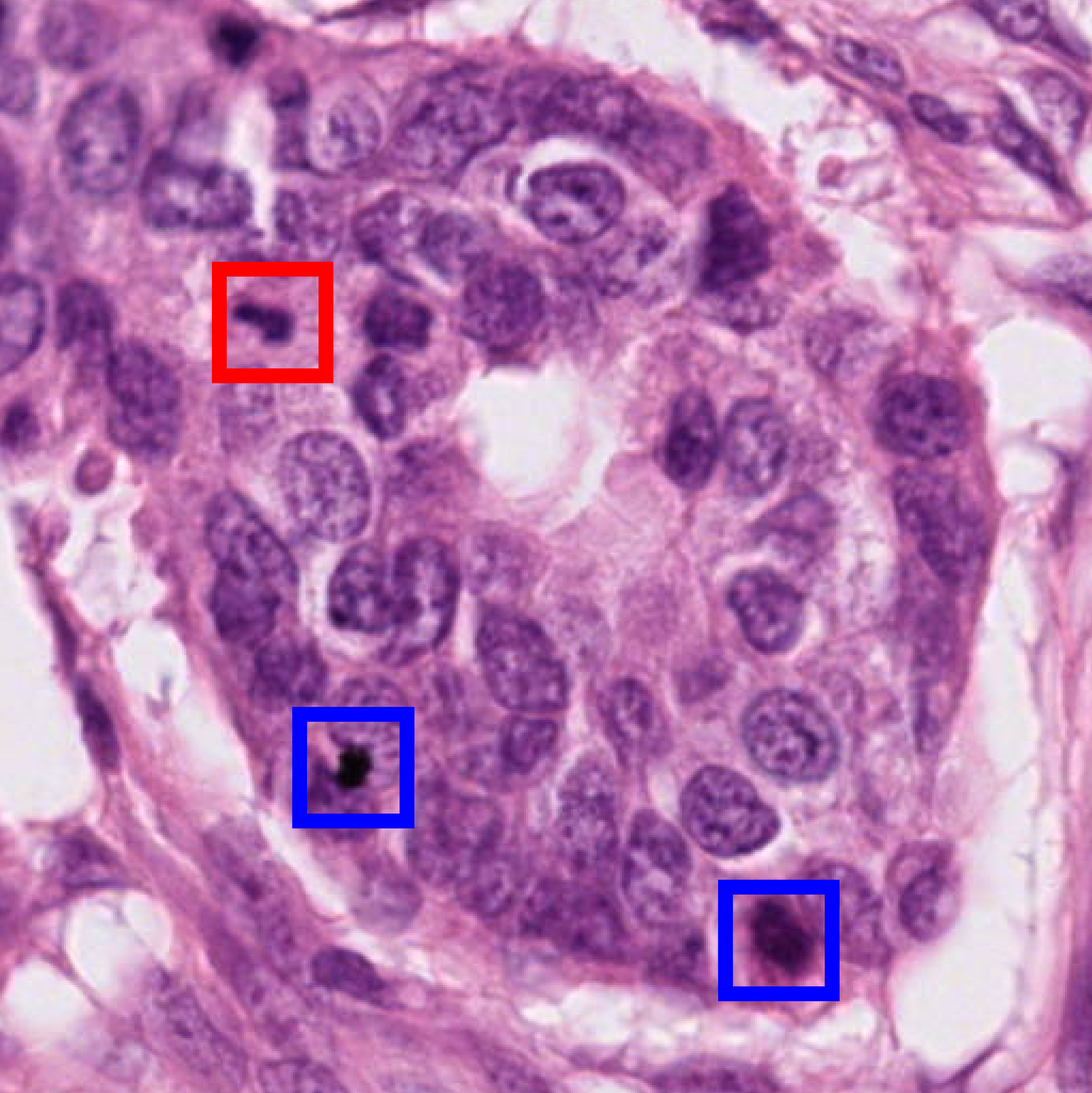}
    \caption{FuseStyle}
    \label{fig:fs}
\end{subfigure}
        
\caption{Mitotic figure detection by different methods in S360 image with model trained on XR \& CS, where Red box$\rightarrow$Mitotic and Blue box$\rightarrow$Non-Mitotic.}
\label{fig:Out}
\end{figure}

\section{Results and Discussion}
\textbf{Classification Task Results:}
We evaluate the state-of-the-art (SOTA) methods along with ours based on their classification performance in out-of-distribution domains, and we use accuracy as the performance metric. Our approach is first compared to the other methods in Table \ref{tab:CAMSOTA}, where the Camelyon17 dataset is used for both training and testing (out-of-distribution). The results presented in Table \ref{tab:CAMSOTA} demonstrate that our approach outperforms all the methods except STRAP \cite{STRAP}.
One should note regarding STRAP that its performance heavily relies on the generated stylized dataset used for training. The time required to generate the stylized data for the Camelyon17- WILDS is around 300 hours in our set up and for the MIDOG’21 Challenge dataset, it is around 75 hours. On the other hand, there is no data generation involved with our FuseStyle. Further, the main operation in FuseStyle is a dot product, which is computationally cheap, and the complexity of our feature mixing strategy is negligible compared to existing augmentation techniques.

Due to the substantial dependence of STRAP on the generated stylized augmentation, careful selection of style images for every dataset becomes fundamental to reproduce its similar performance on different datasets. Therefore, to further investigate the performance of FuseStyle and STRAP, we conduct a classification experiment on the MIDOG'21 Challenge Dataset, the results of which are presented in Table \ref{tab:MDG21}. As seen, if the network is trained on S360 and CS, and tested on XR, there is a 10.23\% advantage in test accuracy for FuseStyle over STRAP. Furthermore, the accuracy improves by 5.79\% and 3.08\% for S360 and CS, which are the seen domains, respectively. In the other cases of Table \ref{tab:MDG21}, the approaches outperform each other almost equal number of times, but most importantly, the differences in their accuracies are relatively low. This shows that FuseStyle is at par with STRAP in these cases in spite of it being significantly less complex. We also infer from the table that FuseStyle produces consistent performance irrespective of the training and testing domains.

\textbf{Mitotic Figure Detection Task Results:}
We conduct an experiment on this task using three different models: Our FuseStyle, STRAP and RetinaNet \cite{retinanet}. All these models use ResNet50 as their backbone architecture, but Retinanet does not involve any domain generalization. We provide Precision, Recall and F1 score as the performance metrics of detection in Table \ref{tab:mfd}. Here the models are trained using training data from XR and CS scanners. As a result, the images from S360 represent an out-of-distribution scenario.
As can be seen, FuseStyle outperforms both STRAP \cite{STRAP} and RetinaNet in most cases in terms of F1 score that incorporates both precision and recall. FuseStyle's superiority over RetinaNet demonstrates the usefulness of our way of domain generalization.

\begin{table}[]
\centering
\resizebox{0.65\textwidth}{!}{%
\begin{tabular}{|l|lll|lll|lll|}
\hline
\multicolumn{1}{|c|}{Network} & \multicolumn{3}{c|}{Precision} & \multicolumn{3}{c|}{Recall} & \multicolumn{3}{c|}{F1 Score} \\ \hline
 & \multicolumn{1}{l|}{XR} & \multicolumn{1}{l|}{S360} & CS & \multicolumn{1}{l|}{XR} & \multicolumn{1}{l|}{S360} & CS & \multicolumn{1}{l|}{XR} & \multicolumn{1}{l|}{S360} & CS \\ \hline
RetinaNet & \multicolumn{1}{l|}{0.91} & \multicolumn{1}{l|}{0.93} & 0.93 & \multicolumn{1}{l|}{0.76} & \multicolumn{1}{l|}{0.3} & 0.76 & \multicolumn{1}{l|}{0.83} & \multicolumn{1}{l|}{0.45} & 0.84 \\ \hline
STRAP & \multicolumn{1}{l|}{0.85} & \multicolumn{1}{l|}{0.91} & 0.88 & \multicolumn{1}{l|}{0.88} & \multicolumn{1}{l|}{0.70} & 0.95 & \multicolumn{1}{l|}{0.87} & \multicolumn{1}{l|}{0.79} & 0.92 \\ \hline
FuseStyle & \multicolumn{1}{l|}{0.82} & \multicolumn{1}{l|}{0.92} & 0.90 & \multicolumn{1}{l|}{0.92} & \multicolumn{1}{l|}{0.76} & 0.90 & \multicolumn{1}{l|}{0.87} & \multicolumn{1}{l|}{0.83} & 0.90 \\ \hline
\end{tabular}%
}
\caption{Mitotic Figure Detection Analysis on $MIDOG'21$ Challenge Dataset.}
\label{tab:mfd}
\end{table}
\vspace{-15pt}
A visual result of mitotic figure detection using FuseStyle, STRAP and RetinaNet is shown in Figure \ref{fig:GT} along with the ground truth. As we can see from the figure, the use of FuseStyle, unlike the use of the other two, results in accurate detection and classification of all mitotic and non-mitotic figures present. While the use of RetinaNet results in an unsuccessful classification of a mitotic figure, the use of STRAP results in detection failure.


\textbf{Design Analysis of Our Approach:}
Our investigation has revealed that combining distant features can lead to the extraction of domain-invariant features. To achieve this, we had proposed using the dot product method, but other techniques for generating a reference batch exist. To explore this further, we conduct an empirical investigation using four different methods: M1: Mixing with Random Shuffle, Reference Approach (RA): Mixing with Least Dot Product (FuseStyle), M2: Mixing with Maximum Euclidean Distance, and M3: Mixing with Maximum KL Divergence. We study the Euclidean distance based approach and also experiment with an advanced approach based on KL divergence. To evaluate the robustness of the proposed approach, we train the ResNet50 model on two scanner datasets and tested it on the third scanner. The comparison of the results obtained from the study are presented in Tables \ref{tab:MIDOG21} \& \ref{tab:MIDOG22}. The comparison is based on the test accuracy (in percentage) of different scanners and the time required for training. The obtained results reveal the effectiveness of the proposed approach of sample selection for mixing. The detailed analysis of the findings is provided in the table, demonstrating the superiority of the proposed method over the other methods.
\begin{center}
\begin{minipage}[t]{0.99\textwidth}
\centering
  \begin{minipage}[t]{0.49\textwidth}
  \captionof{table}{Objective evaluation on $MIDOG'21$  Challenge Dataset.}
  \resizebox{\textwidth}{!}{
  \centering
\begin{tabular}{|c|c|c|c|c|c|}
\hline
Network                                                                                     & Methods & XR             & S360           & CS             & \begin{tabular}[c]{@{}c@{}}Time/epoch\\ (sec)\end{tabular} \\ \hline
\multirow{4}{*}{\begin{tabular}[c]{@{}c@{}}Train: \\ S360 \& CS\\ Test: \\ XR\end{tabular}} & M1      & 76.42          & 89.45          & 88.76          & 08                                                         \\ \cline{2-6} 
                                                                                            & RA      & \textbf{77.56} & \textbf{91.07} & \textbf{90.46} & 08                                                         \\ \cline{2-6} 
                                                                                            & M2      & 67.05          & 88.64          & 90.29          & 08                                                         \\ \cline{2-6} 
                                                                                            & M3      & 74.43          & 87.18          & 88.76          & 111                                                        \\ \hline
\multirow{4}{*}{\begin{tabular}[c]{@{}c@{}}Train:\\ XR \& CS\\ Test: \\ S360\end{tabular}}  & M1      & 84.09          & 71.59          & 85.18          & 08                                                         \\ \cline{2-6} 
                                                                                            & RA      & 90.06          & 75.16          & \textbf{92.16} & 08                                                         \\ \cline{2-6} 
                                                                                            & M2      & \textbf{90.34} & 75.65          & 91.65          & 08                                                         \\ \cline{2-6} 
                                                                                            & M3      & 83.07          & \textbf{78.90} & 89.95          & 111                                                        \\ \hline
\multirow{4}{*}{\begin{tabular}[c]{@{}c@{}}Train:\\ XR \& S360\\ Test: \\ CS\end{tabular}}  & M1      & \textbf{88.64} & 89.45          & 72.40          & 08                                                         \\ \cline{2-6} 
                                                                                            & RA      & 86.65          & 88.96          & 74.10          & 08                                                         \\ \cline{2-6} 
                                                                                            & M2      & 87.78          & \textbf{90.91} & 76.32          & 08                                                         \\ \cline{2-6} 
                                                                                            & M3      & 86.34          & 87.34          & \textbf{78.02} & 111                                                        \\ \hline
\end{tabular}}
     \label{tab:MIDOG21}
  \end{minipage}
  \hfill
  \begin{minipage}[t]{0.49\textwidth}
      \captionof{table}{Objective evaluation on  $MIDOG'22$ Challenge Dataset.}
  \resizebox{\textwidth}{!}{
    \centering
\begin{tabular}{|c|c|c|c|c|c|}
\hline
Network                                                                                     & Methods & XR             & S360           & CS             & \begin{tabular}[c]{@{}c@{}}Time/epoch\\ (sec)\end{tabular} \\ \hline
\multirow{4}{*}{\begin{tabular}[c]{@{}c@{}}Train: \\ S360 \& CS\\ Test: \\ XR\end{tabular}} & M1      & 75.10          & 81.16          & 80.58          & 28                                                         \\ \cline{2-6} 
                                                                                            & RA      & 74.27          & \textbf{83.59} & 79.74          & 28                                                         \\ \cline{2-6} 
                                                                                            & M2      & 76.78          & 75.68          & 68.80          & 28                                                         \\ \cline{2-6} 
                                                                                            & M3      & \textbf{78.74} & 83.28          & \textbf{81.24} & 331                                                        \\ \hline
\multirow{4}{*}{\begin{tabular}[c]{@{}c@{}}Train:\\ XR \& CS\\ Test: \\ S360\end{tabular}}  & M1      & 77.06          & 80.24          & 78.89          & 28                                                         \\ \cline{2-6} 
                                                                                            & RA      & 80.84          & 80.24          & \textbf{81.62} & 28                                                         \\ \cline{2-6} 
                                                                                            & M2      & 75.94          & \textbf{81.76} & 81.53          & 28                                                         \\ \cline{2-6} 
                                                                                            & M3      & \textbf{81.26} & \textbf{81.76} & 79.26          & 331                                                        \\ \hline
\multirow{4}{*}{\begin{tabular}[c]{@{}c@{}}Train:\\ XR \& S360\\ Test: \\ CS\end{tabular}}  & M1      & 81.12          & 81.46          & 82.28          & 28                                                         \\ \cline{2-6} 
                                                                                            & RA      & \textbf{81.34} & \textbf{84.19} & \textbf{82.75} & 28                                                         \\ \cline{2-6} 
                                                                                            & M2      & 73.71          & 76.60          & 73.42          & 28                                                         \\ \cline{2-6} 
                                                                                            & M3      & 80.14          & 83.89          & 74.08          & 331                                                        \\ \hline
\end{tabular}}
    \label{tab:MIDOG22}
  \end{minipage}
\end{minipage}
\end{center}
\vspace{0.5cm}
Based on the results presented in Table \ref{tab:MIDOG21}, it can be observed that the Dot Product method is the most consistent in terms of network performance across different domains. In contrast, the Random Shuffle method (M1) fails to perform well in the second case, and the Euclidean Distance method (M2) fails in the first case for the out-of-distribution domain. The KL Divergence method (M3) does not perform well in the in-distribution domain, as observed in the second case for the XR scanner, and it also requires a significantly longer computational time compared to the other methods. Therefore, the experimental studies suggest that FuseStyle (RA) provides the most consistent results as well as takes less time compared to KL divergence method (M3) on the MIDOG'21 Challenge dataset. For further analysis of our method, we conduct additional experiments on the MIDOG'22 Challenge dataset \cite{midog22} as shown in Table \ref{tab:MIDOG22}, using the same reference batch generation methods. The results demonstrate that the FuseStyle (RA) performs well for both in-distribution and out-of-distribution domains.

\section{Conclusion}
We present, FuseStyle, a novel method that computes generalized features by mixing them in the feature space to address domain shift issues related to histopathological images. It uses a new approach of feature mixing based on correlation computation. FuseStyle has lower computational requirements, with dot product being the main operation in it. We have shown that the performance of our method in classification and detection tasks is at par or better than the state-of-the-art on various datasets. We also find from experimental results that the proposed feature-mixing method has strong domain generalization capabilities. In summary, our method is simple, effective and consistent, and it has the potential to enhance the out-of-distribution performance of any existing machine learning method.


 \bibliographystyle{splncs04}
 \bibliography{ref}

\end{document}


\fontsize{15}{30}{\textbf{Visual Result of Mitotic Figure Detection}}\\
\section{Train: XR and CS, Test: S360}
\subsection{Unseen Domain: S360}
\vspace{-0.75cm}
\begin{figure}[H]
\centering
\begin{subfigure}{0.24\textwidth}
    \includegraphics[width=\textwidth]{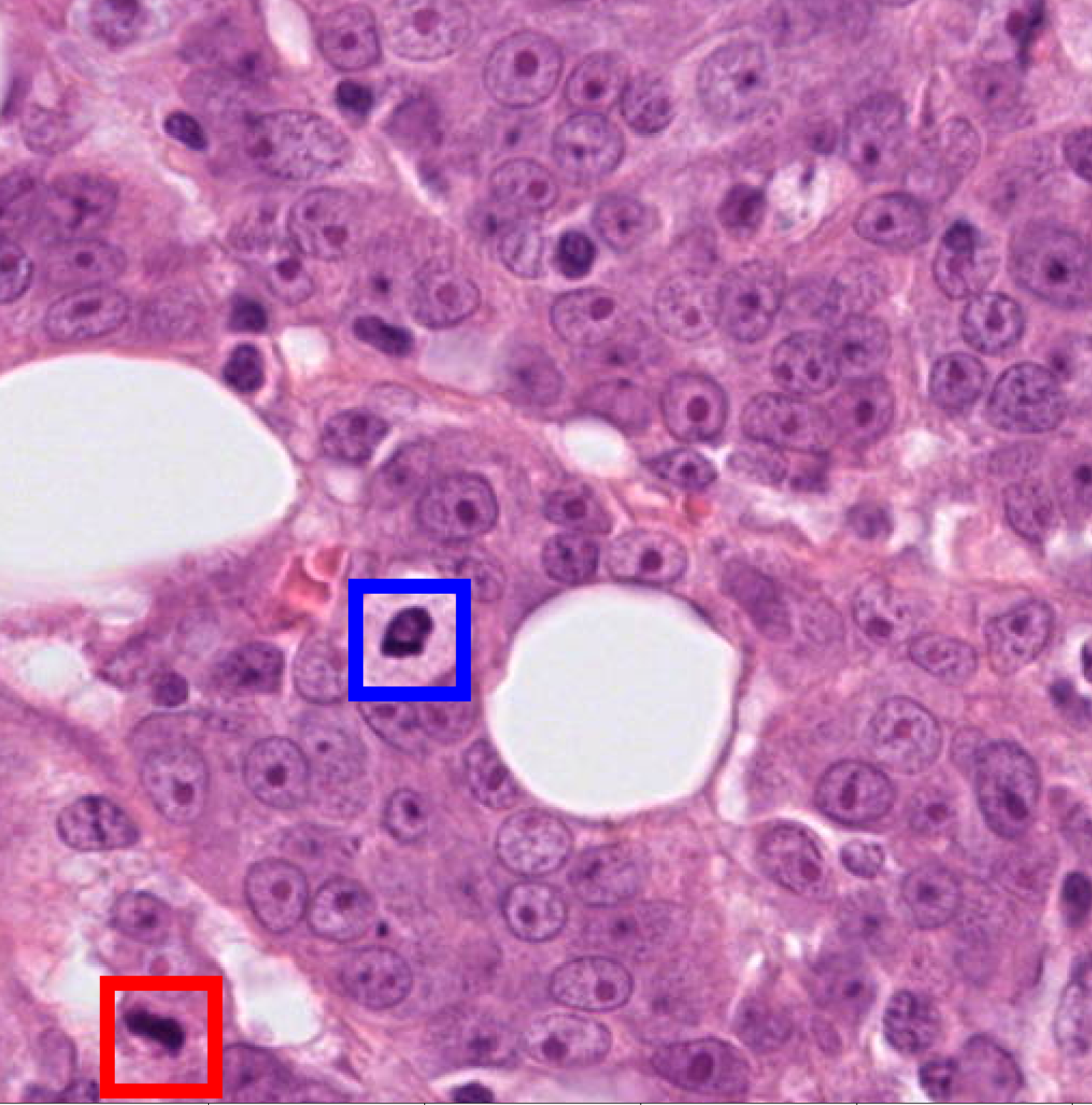}
    \caption{Ground Truth}
    \label{fig:GT}
\end{subfigure}
\hfill
\begin{subfigure}{0.24\textwidth}
    \includegraphics[width=\textwidth]{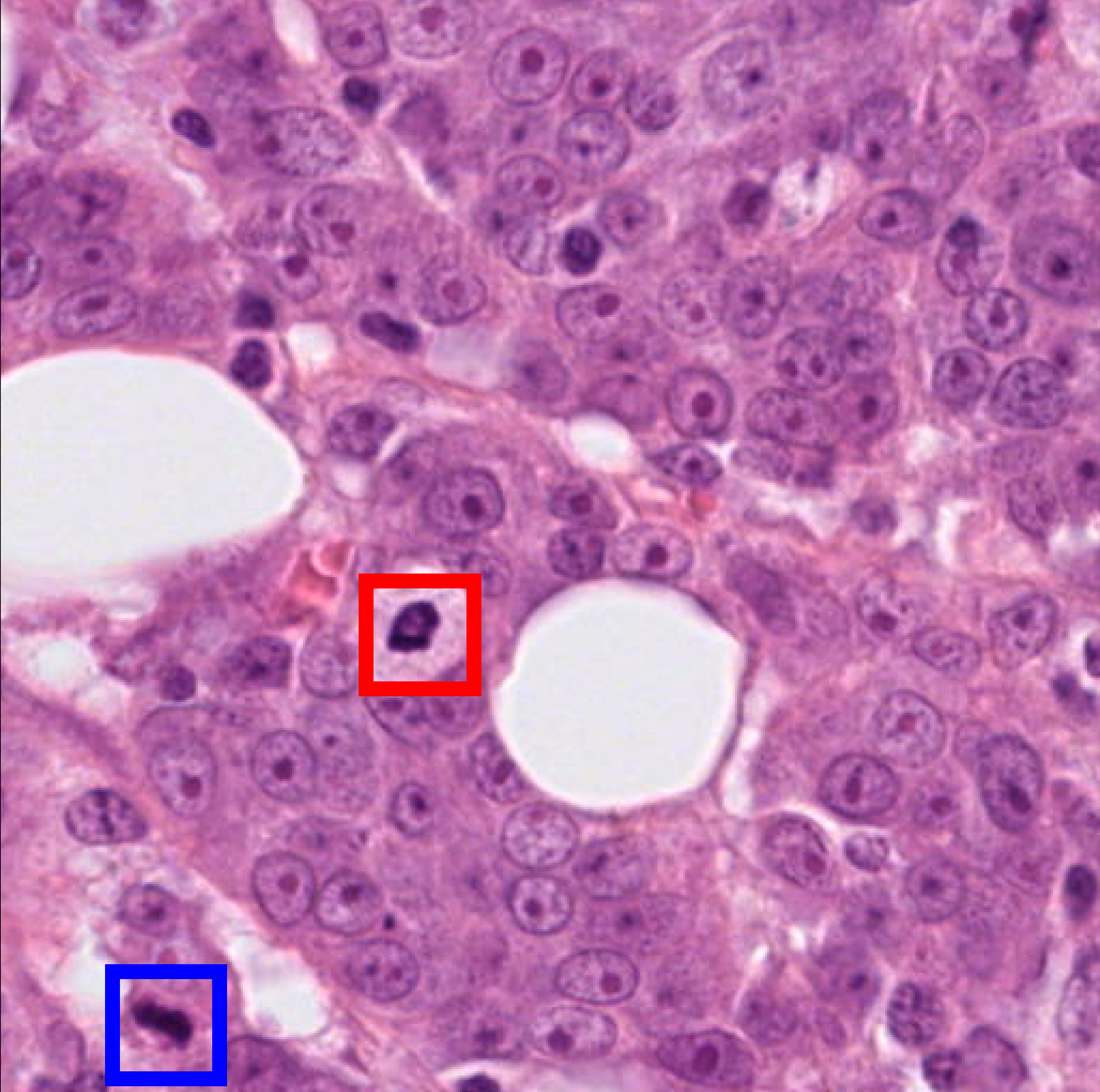}
    \caption{RetinaNet}
    \label{fig:WF}
\end{subfigure}
\begin{subfigure}{0.24\textwidth}
    \includegraphics[width=\textwidth]{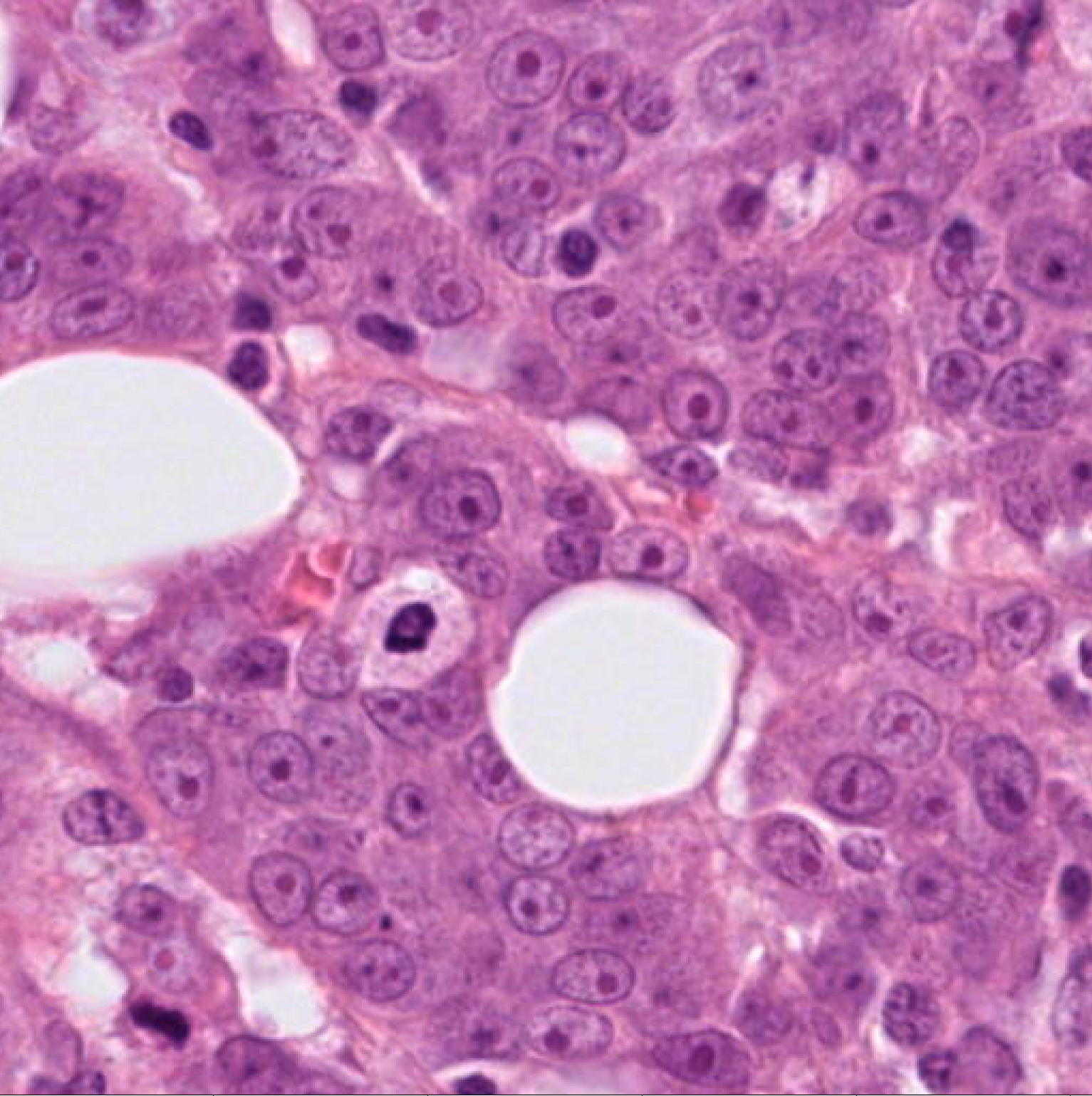}
    \caption{STRAP}
    \label{fig:strap}
\end{subfigure}
\begin{subfigure}{0.24\textwidth}
    \includegraphics[width=\textwidth]{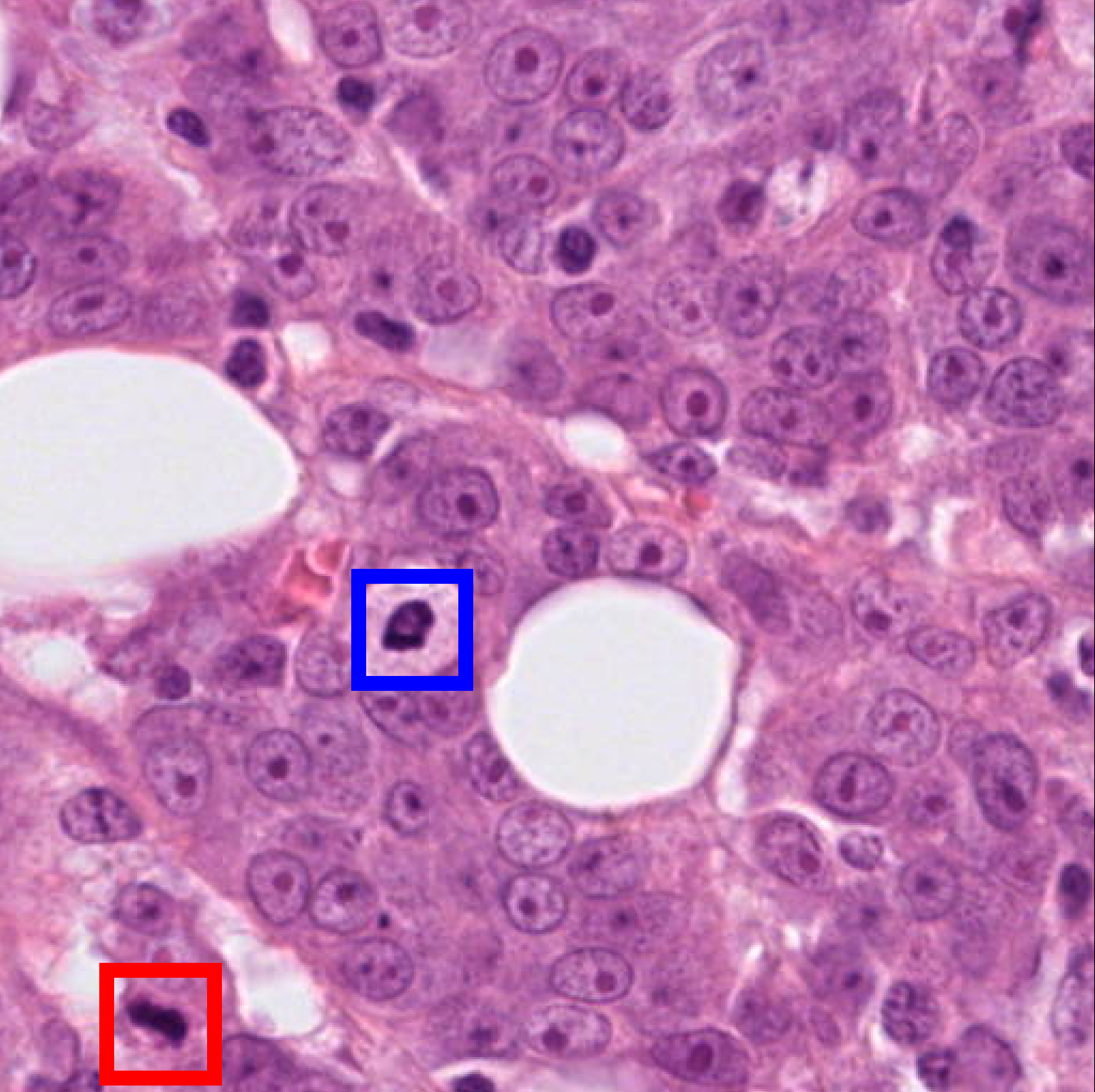}
    \caption{FuseStyle}
    \label{fig:fs}
\end{subfigure}
\label{fig:Out}
\end{figure}

\vspace{-0.75cm}
\subsection{Seen Domain: XR}
\vspace{-0.75cm}
\begin{figure}[H]
\centering
\begin{subfigure}{0.24\textwidth}
    \includegraphics[width=\textwidth]{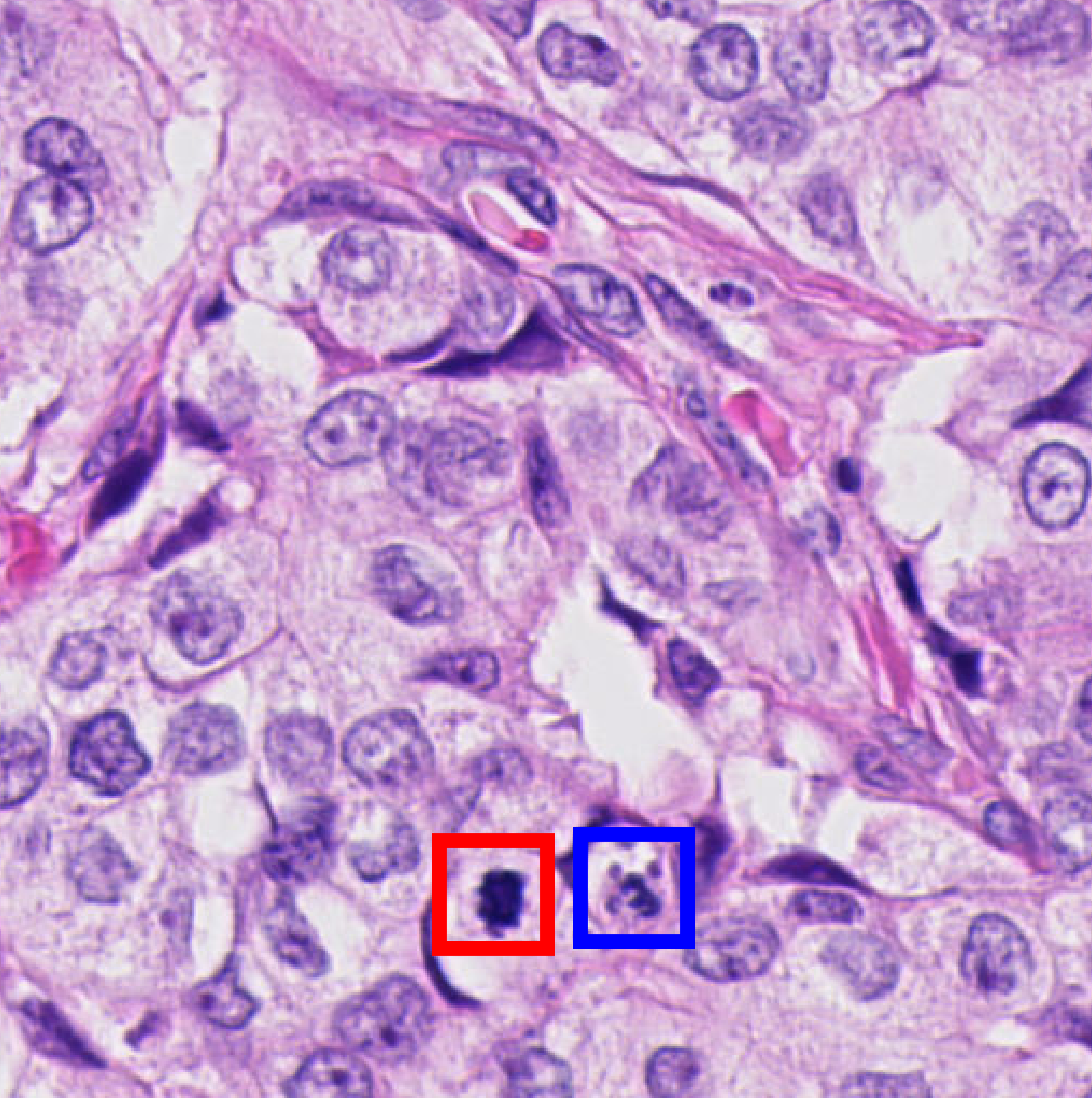}
    \caption{Ground Truth}
    \label{fig:GT}
\end{subfigure}
\hfill
\begin{subfigure}{0.24\textwidth}
    \includegraphics[width=\textwidth]{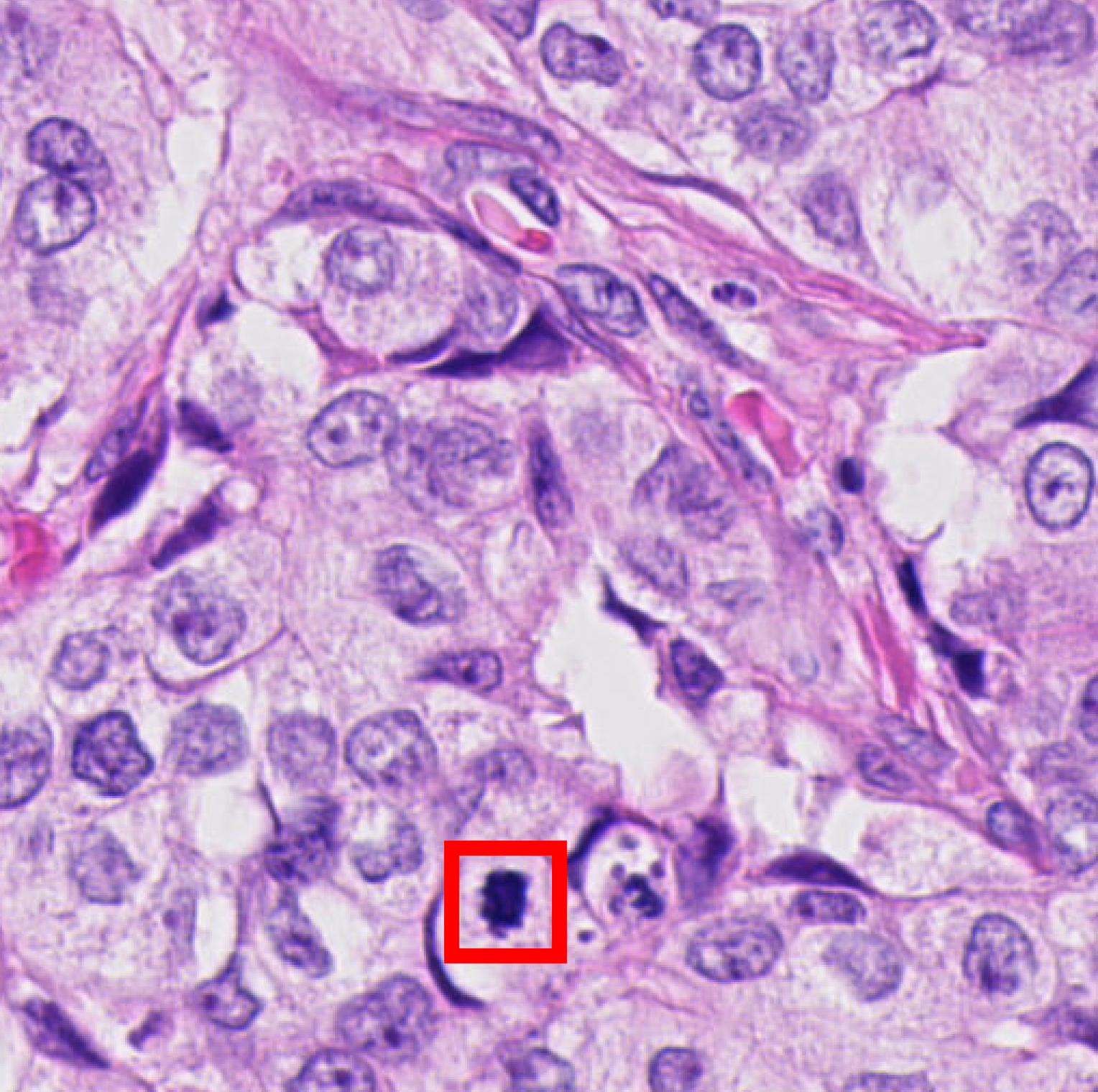}
    \caption{RetinaNet}
    \label{fig:WF}
\end{subfigure}
\hfill
\begin{subfigure}{0.24\textwidth}
    \includegraphics[width=\textwidth]{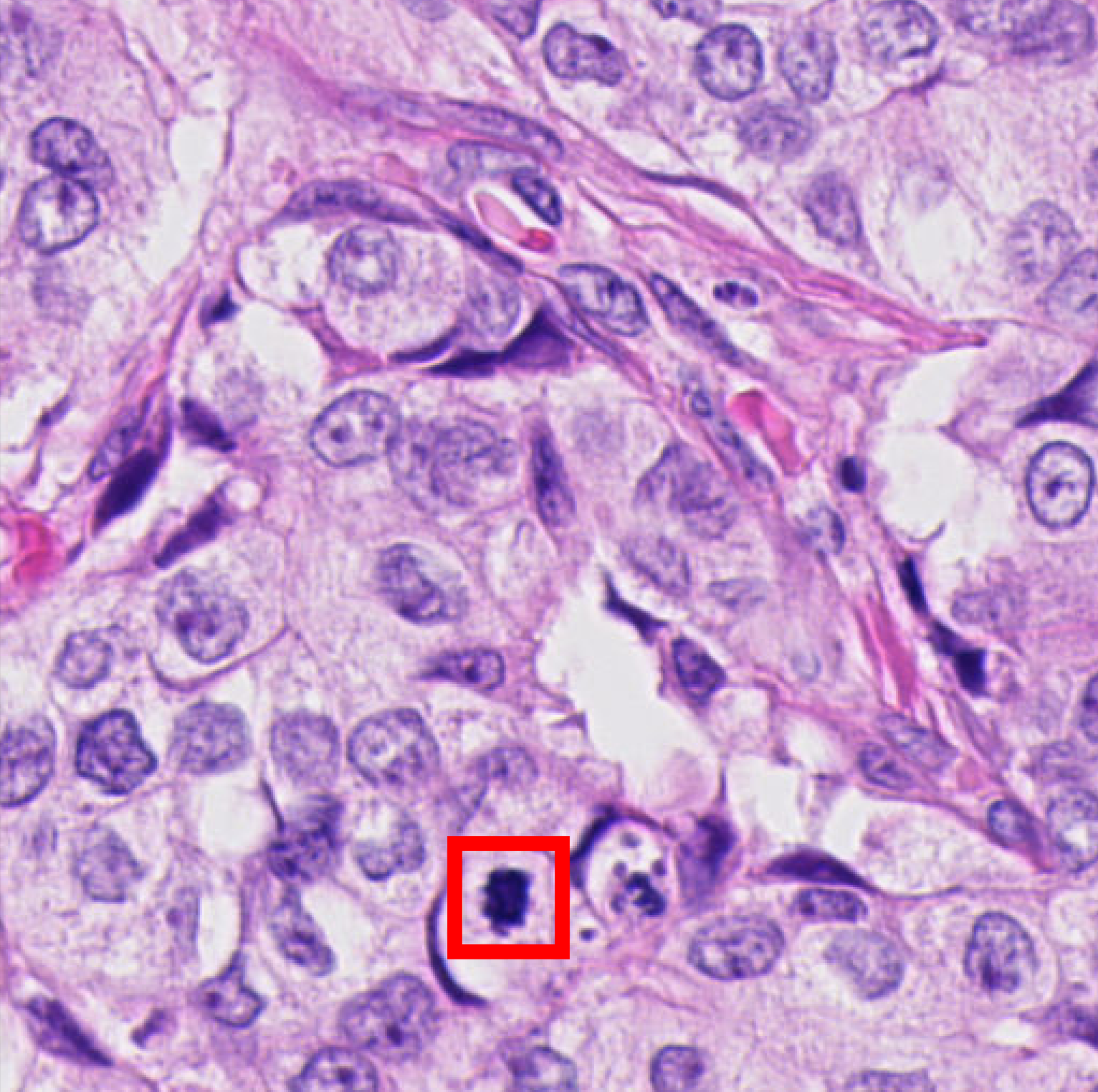}
    \caption{STRAP}
    \label{fig:strap}
\end{subfigure}
\hfill
\begin{subfigure}{0.24\textwidth}
    \includegraphics[width=\textwidth]{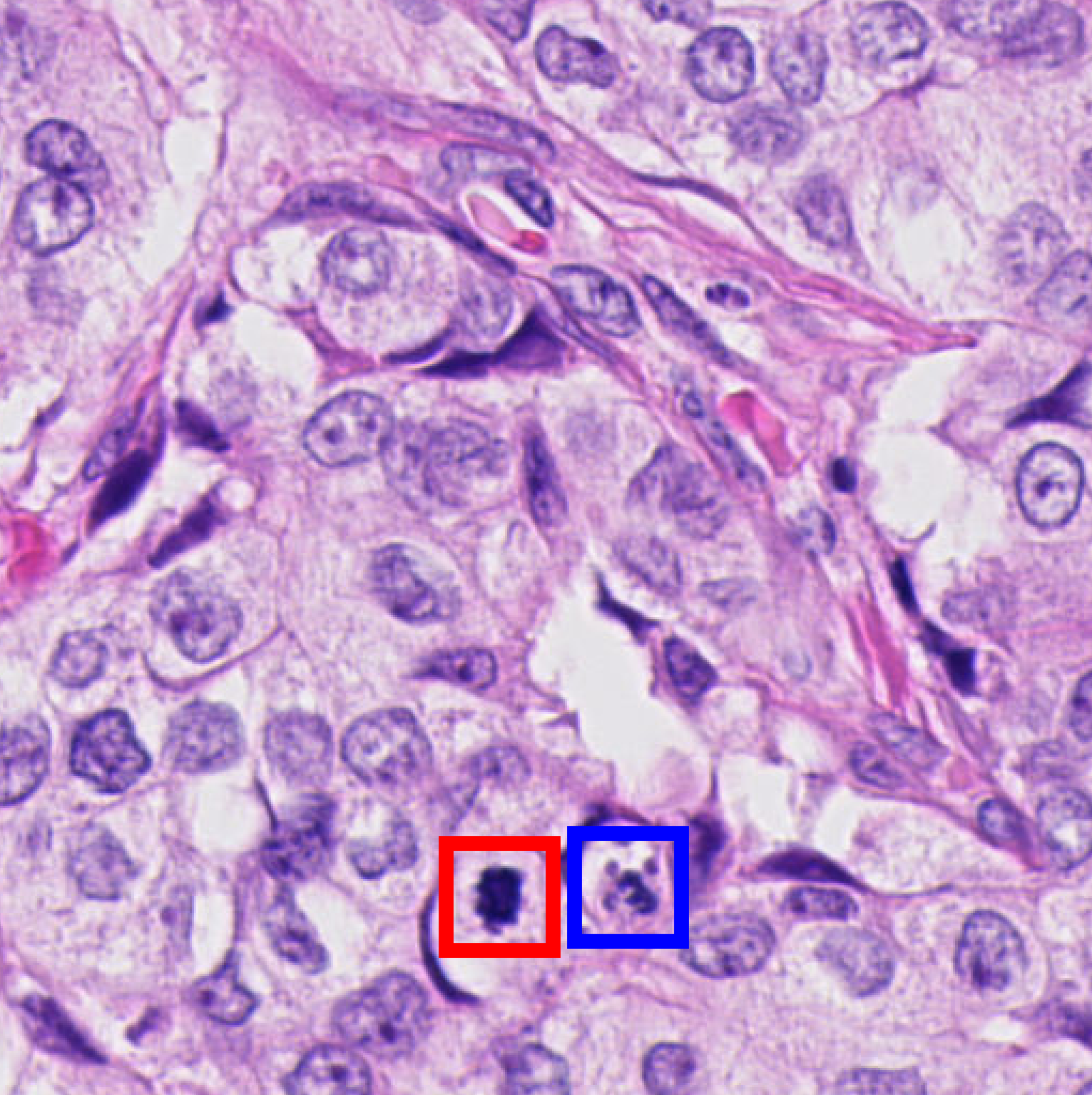}
    \caption{FuseStyle}
    \label{fig:fs}
\end{subfigure}
\label{fig:Out}
\end{figure}

\vspace{-0.75cm}
\subsection{Seen Domain: CS}
\vspace{-0.75cm}
\begin{figure}[H]
\centering
\begin{subfigure}{0.24\textwidth}
    \includegraphics[width=\textwidth]{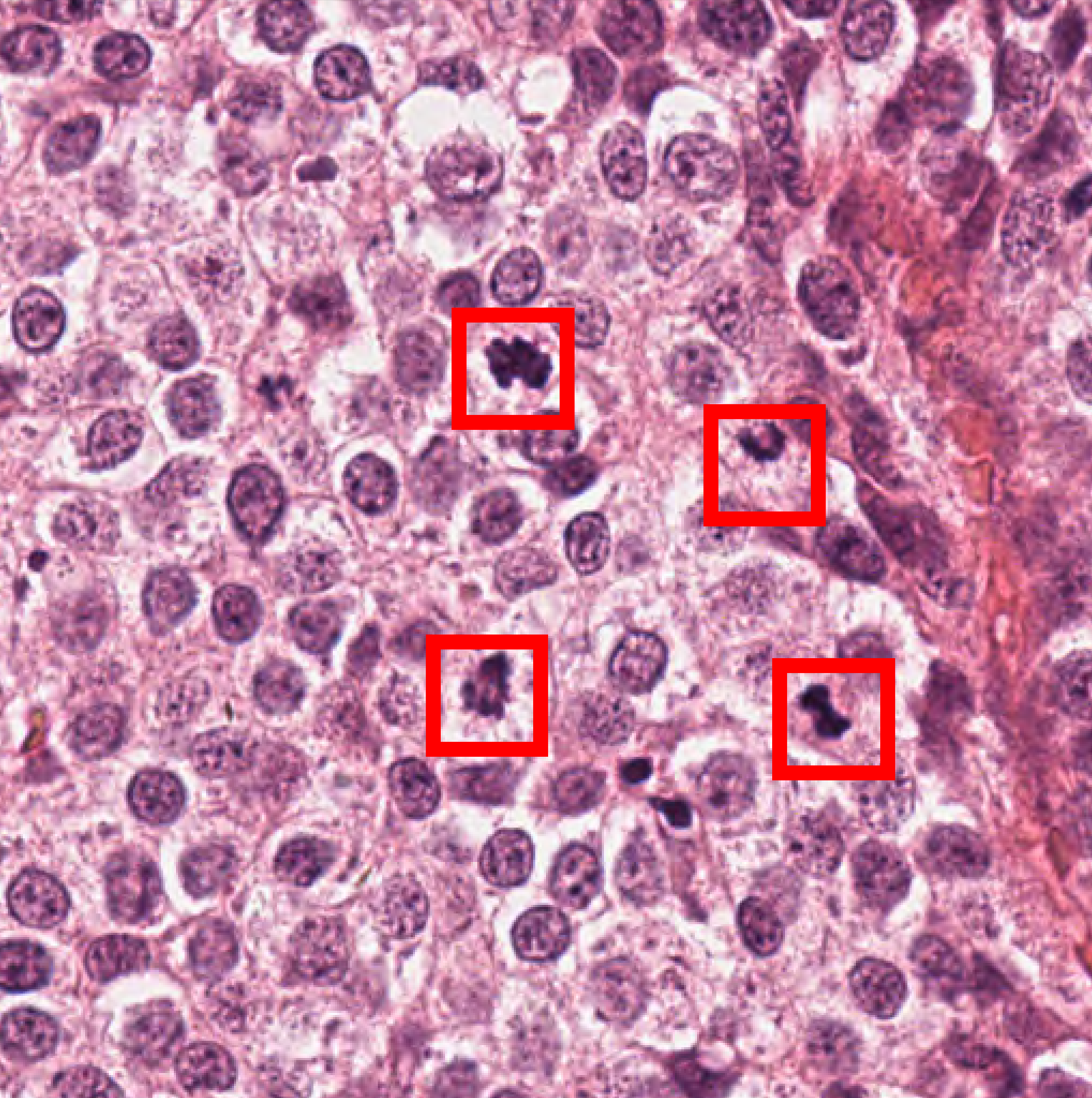}
    \caption{Ground Truth}
    \label{fig:GT}
\end{subfigure}
\hfill
\begin{subfigure}{0.24\textwidth}
    \includegraphics[width=\textwidth]{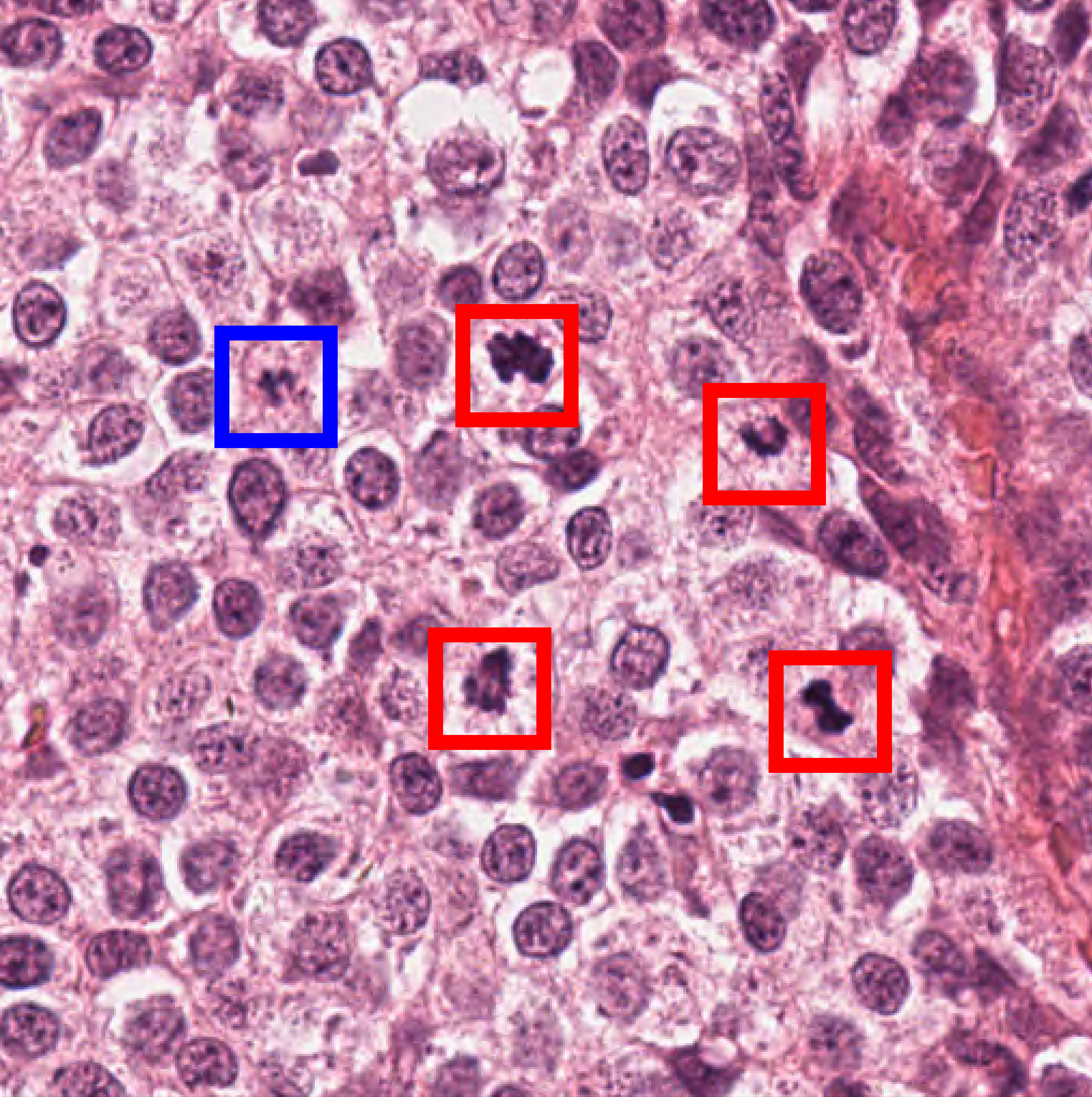}
    \caption{RetinaNet}
    \label{fig:WF}
\end{subfigure}
\hfill
\begin{subfigure}{0.24\textwidth}
    \includegraphics[width=\textwidth]{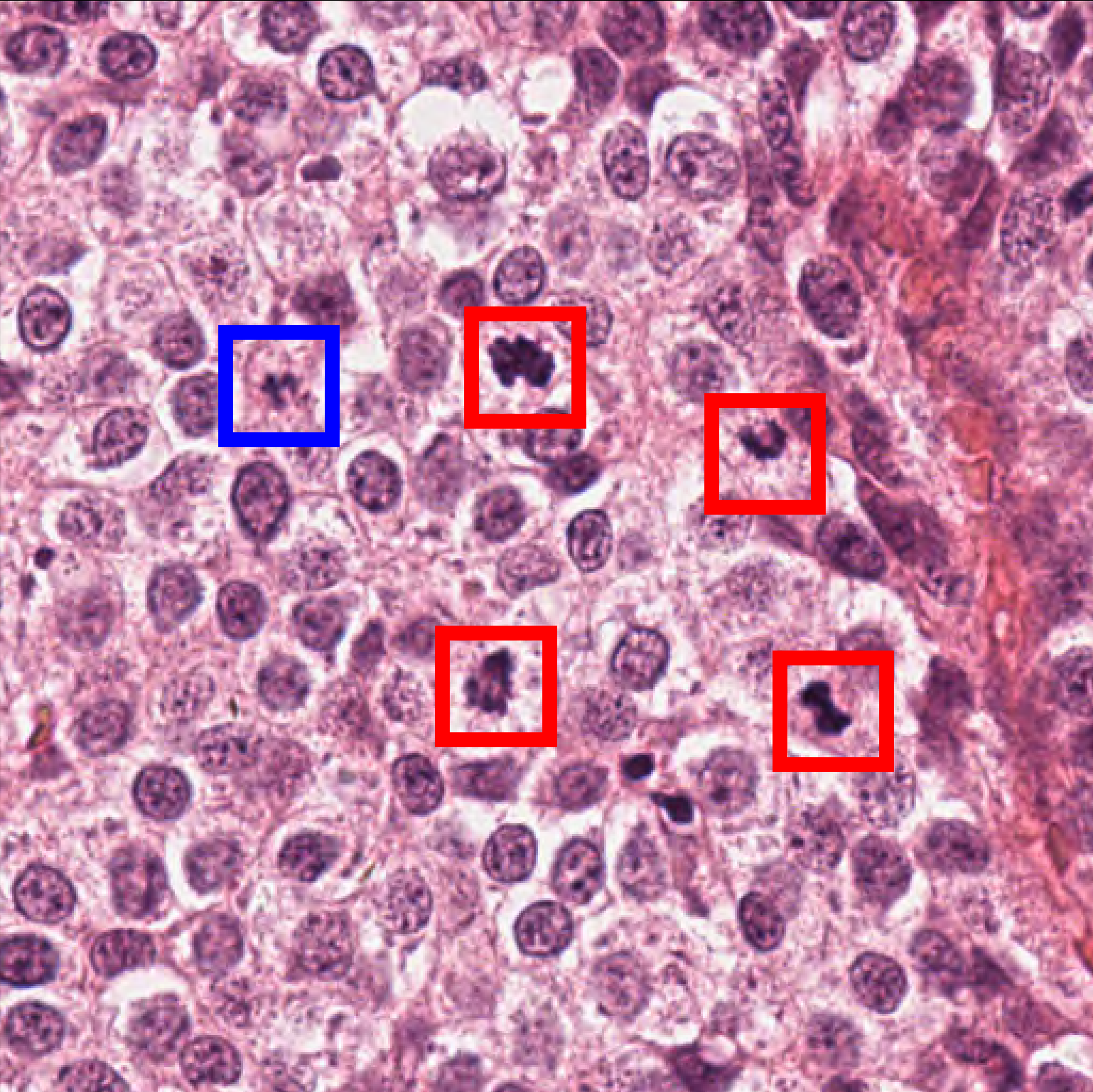}
    \caption{STRAP}
    \label{fig:strap}
\end{subfigure}
\hfill
\begin{subfigure}{0.24\textwidth}
    \includegraphics[width=\textwidth]{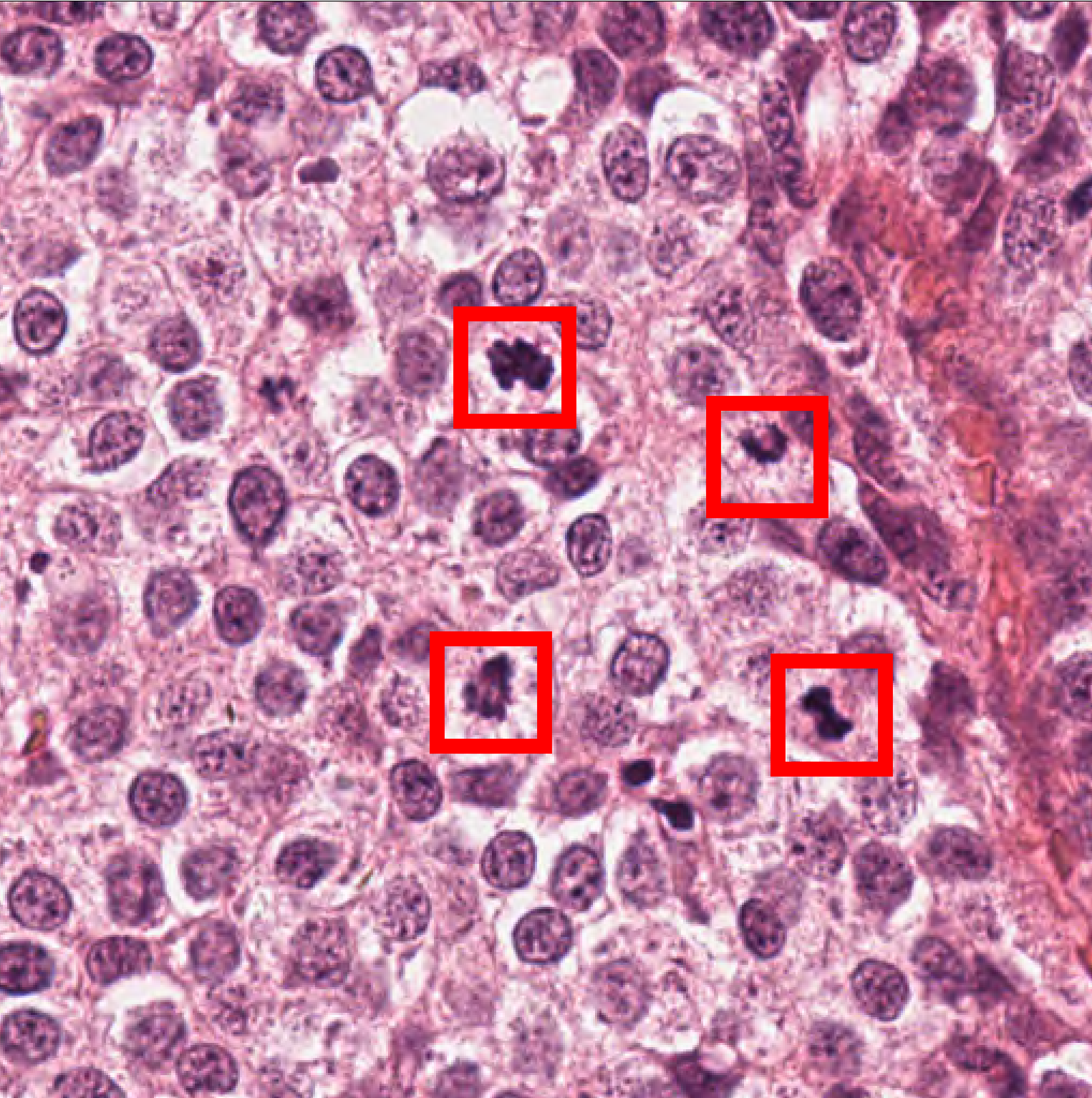}
    \caption{FuseStyle}
    \label{fig:fs}
\end{subfigure}
\label{fig:Out}
\end{figure}
\vspace{-0.75cm}

\section{Train: XR and S360, Test: CS}
\subsection{Unseen Domain: CS}
\vspace{-0.75cm}
\begin{figure}[H]
\centering
\begin{subfigure}{0.24\textwidth}
    \includegraphics[width=\textwidth]{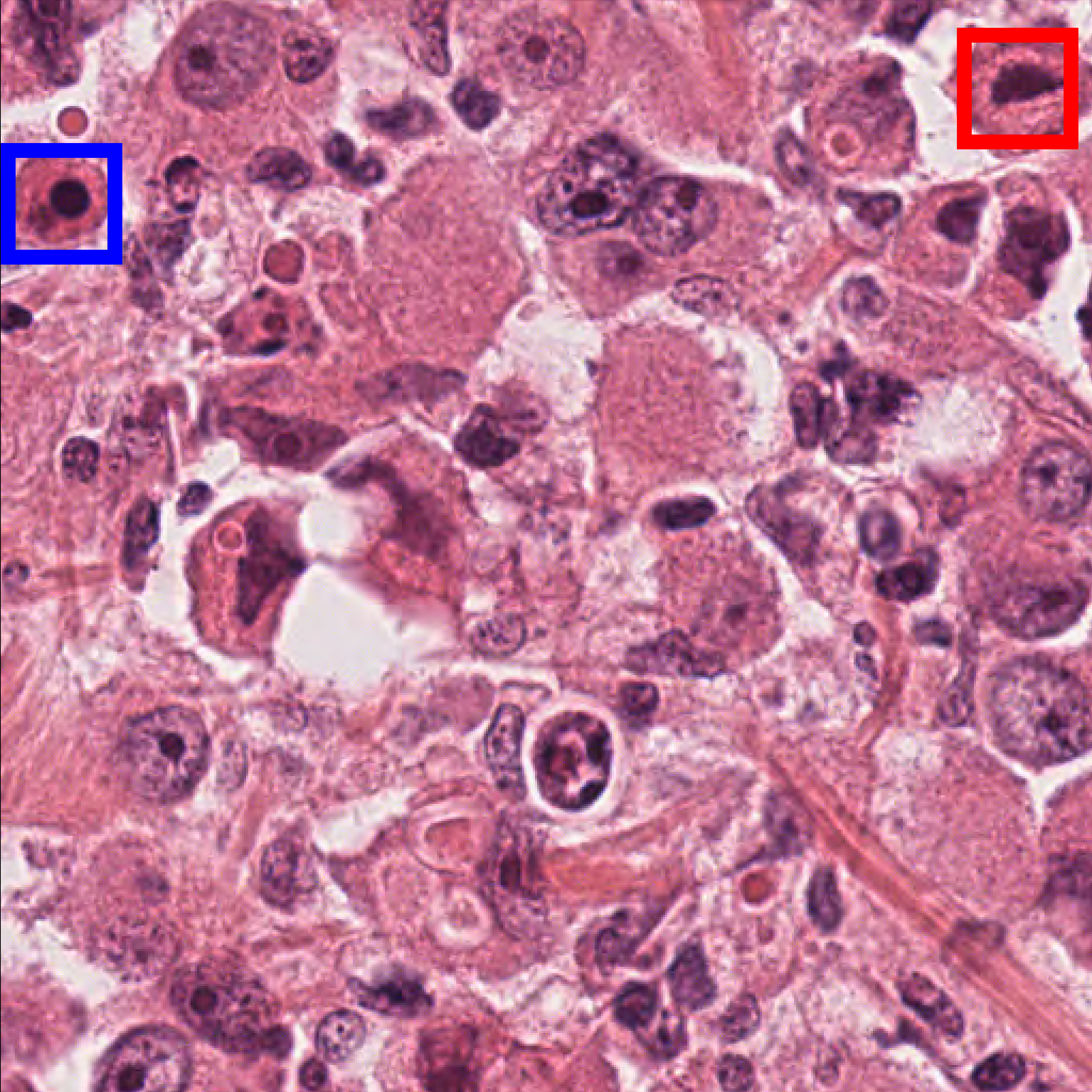}
    \caption{Ground Truth}
    \label{fig:GT}
\end{subfigure}
\hfill
\begin{subfigure}{0.24\textwidth}
    \includegraphics[width=\textwidth]{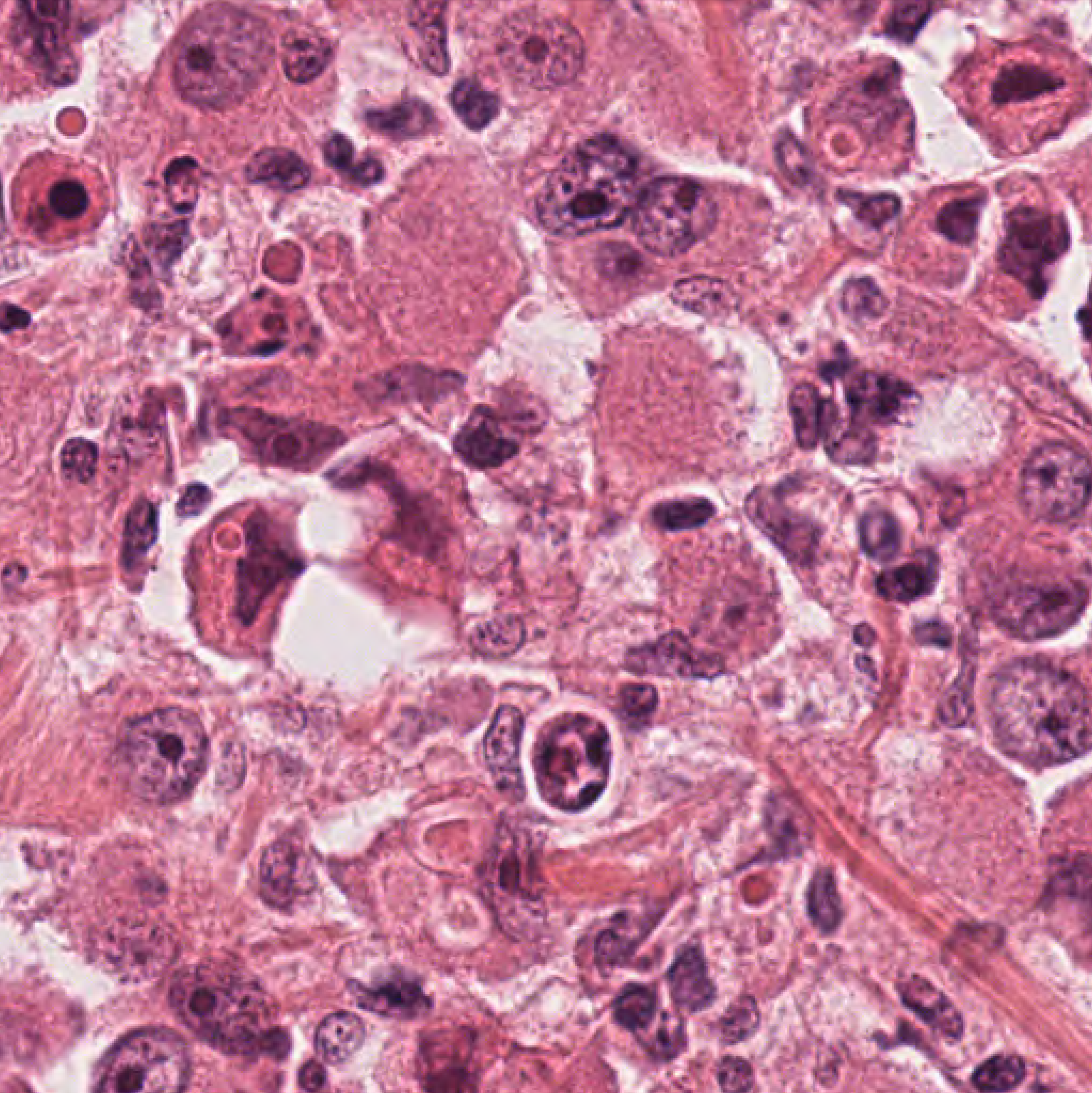}
    \caption{RetinaNet}
    \label{fig:WF}
\end{subfigure}
\hfill
\begin{subfigure}{0.24\textwidth}
    \includegraphics[width=\textwidth]{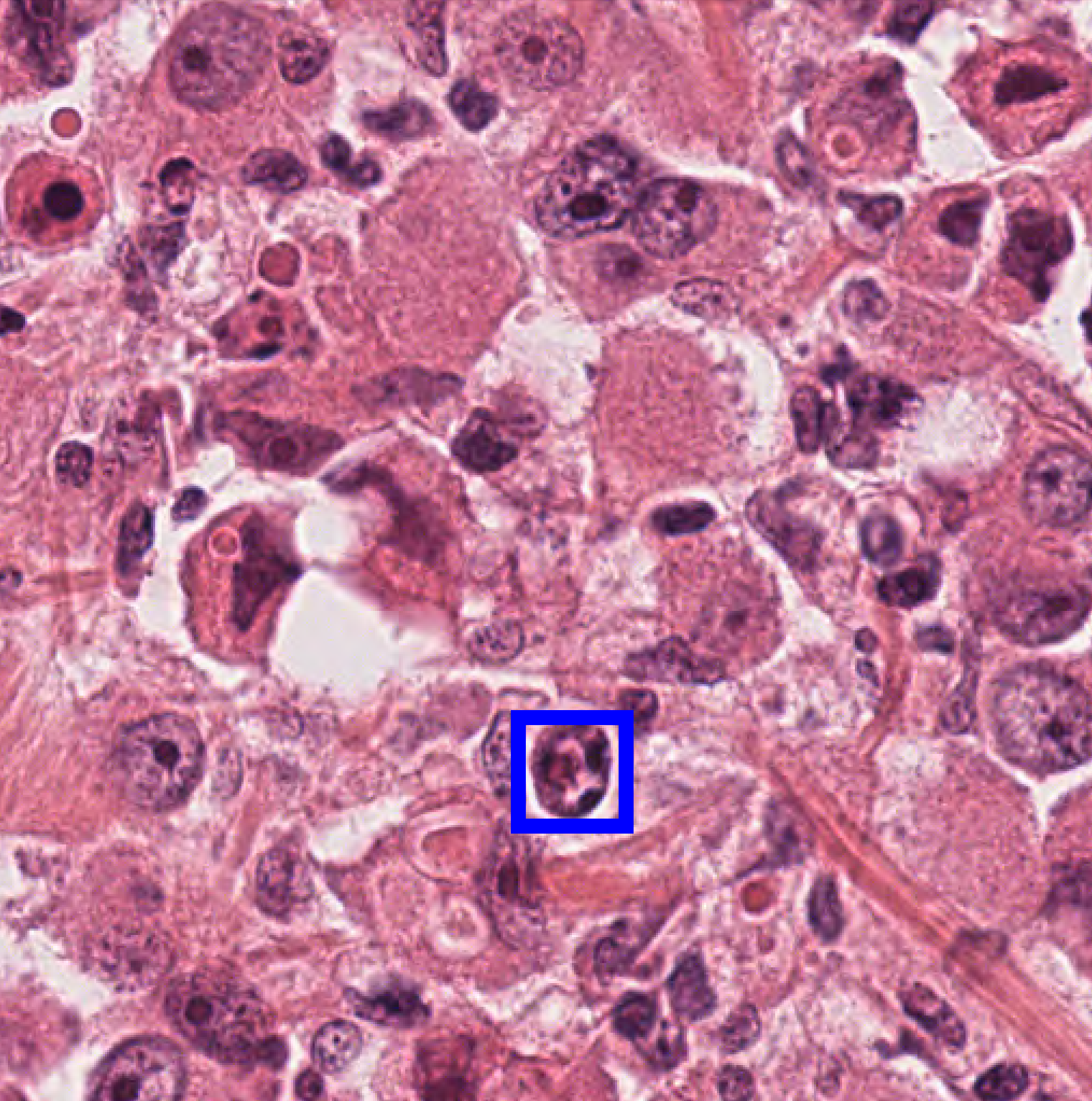}
    \caption{STRAP}
    \label{fig:strap}
\end{subfigure}
\hfill
\begin{subfigure}{0.24\textwidth}
    \includegraphics[width=\textwidth]{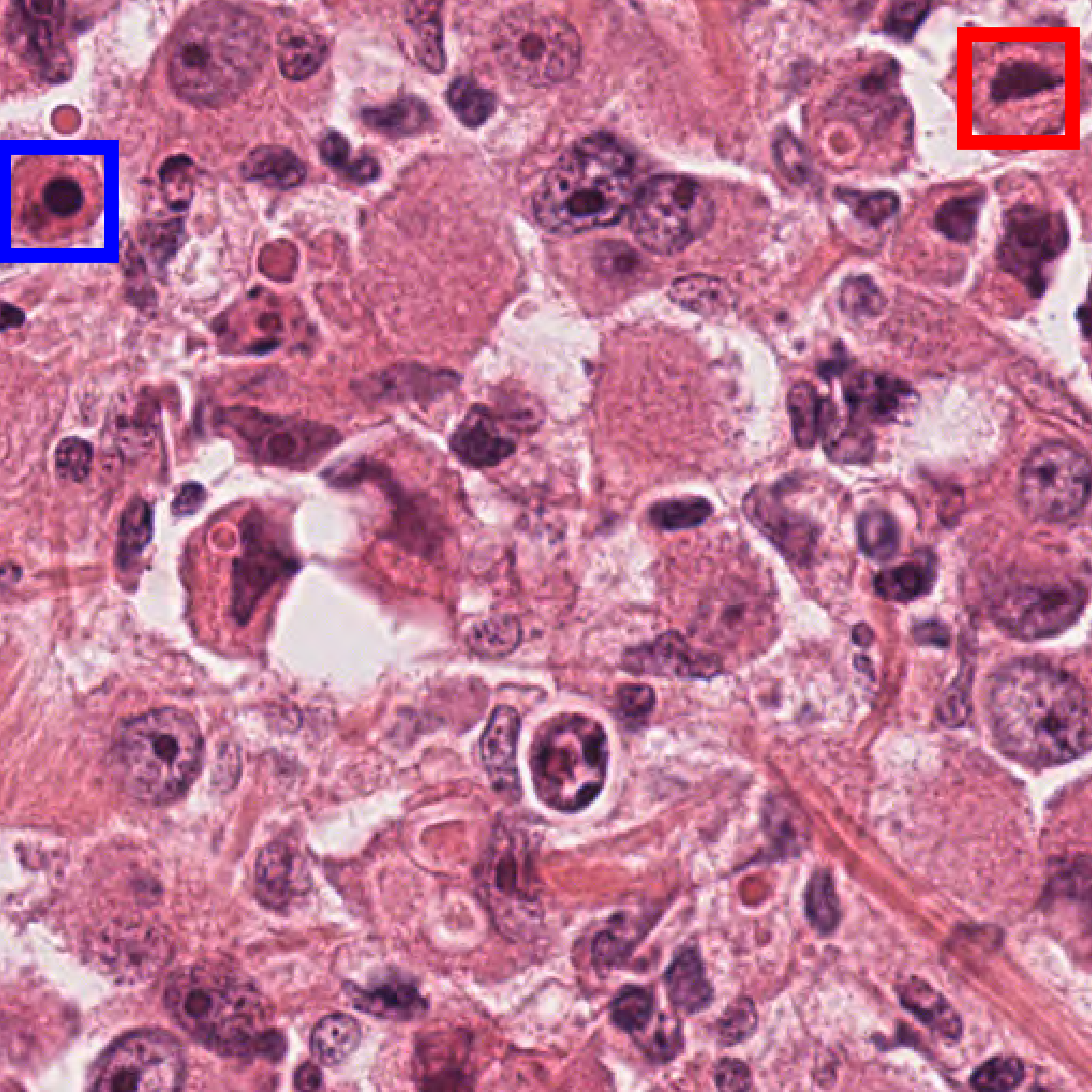}
    \caption{FuseStyle}
    \label{fig:fs}
\end{subfigure}
\label{fig:Out}
\end{figure}

\vspace{-0.75cm}
\subsection{Seen Domain: XR}
\vspace{-0.75cm}
\begin{figure}[H]
\centering
\begin{subfigure}{0.24\textwidth}
    \includegraphics[width=\textwidth]{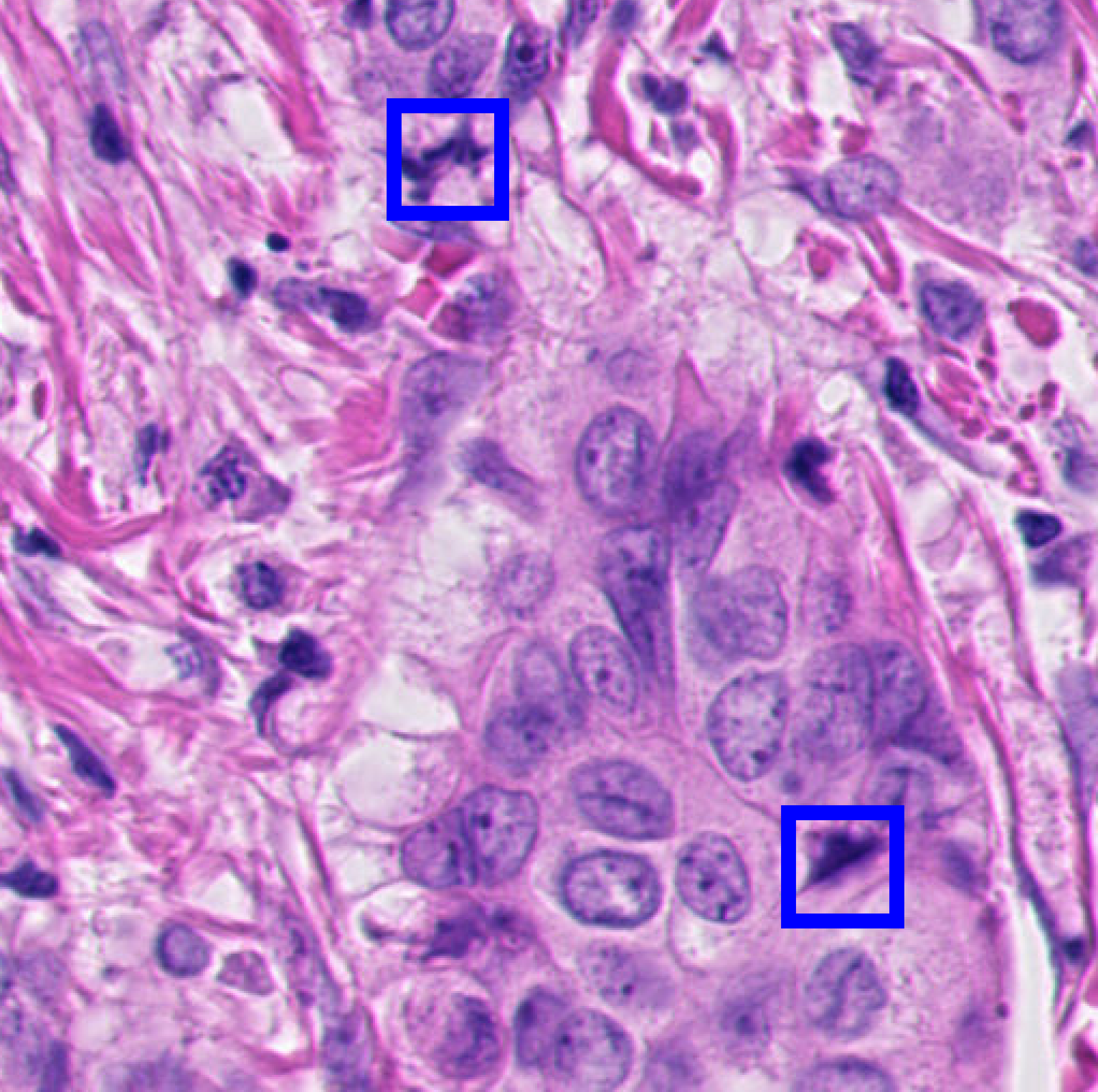}
    \caption{Ground Truth}
    \label{fig:GT}
\end{subfigure}
\begin{subfigure}{0.24\textwidth}
    \includegraphics[width=\textwidth]{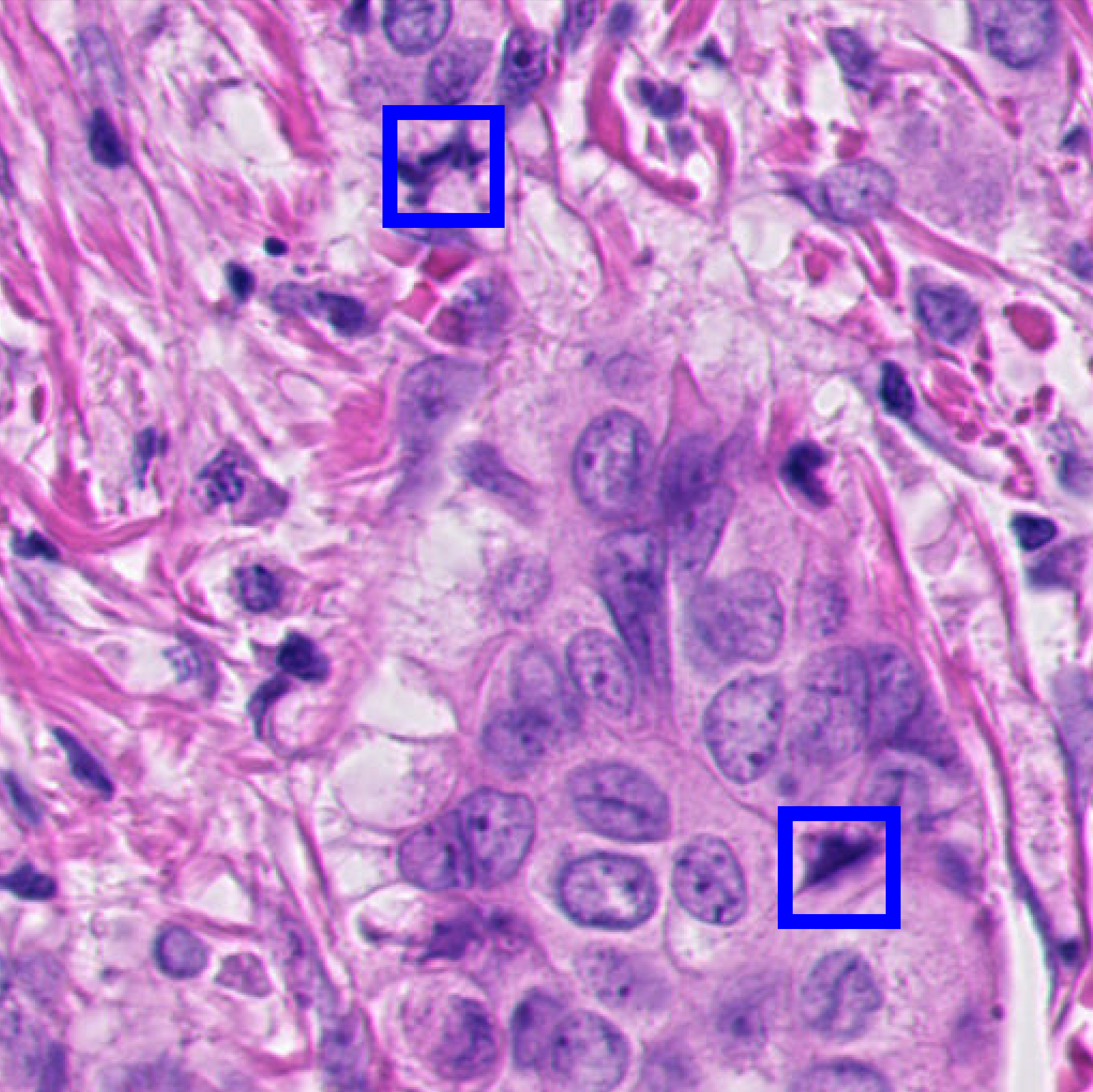}
    \caption{RetinaNet}
    \label{fig:WF}
\end{subfigure}
\begin{subfigure}{0.24\textwidth}
    \includegraphics[width=\textwidth]{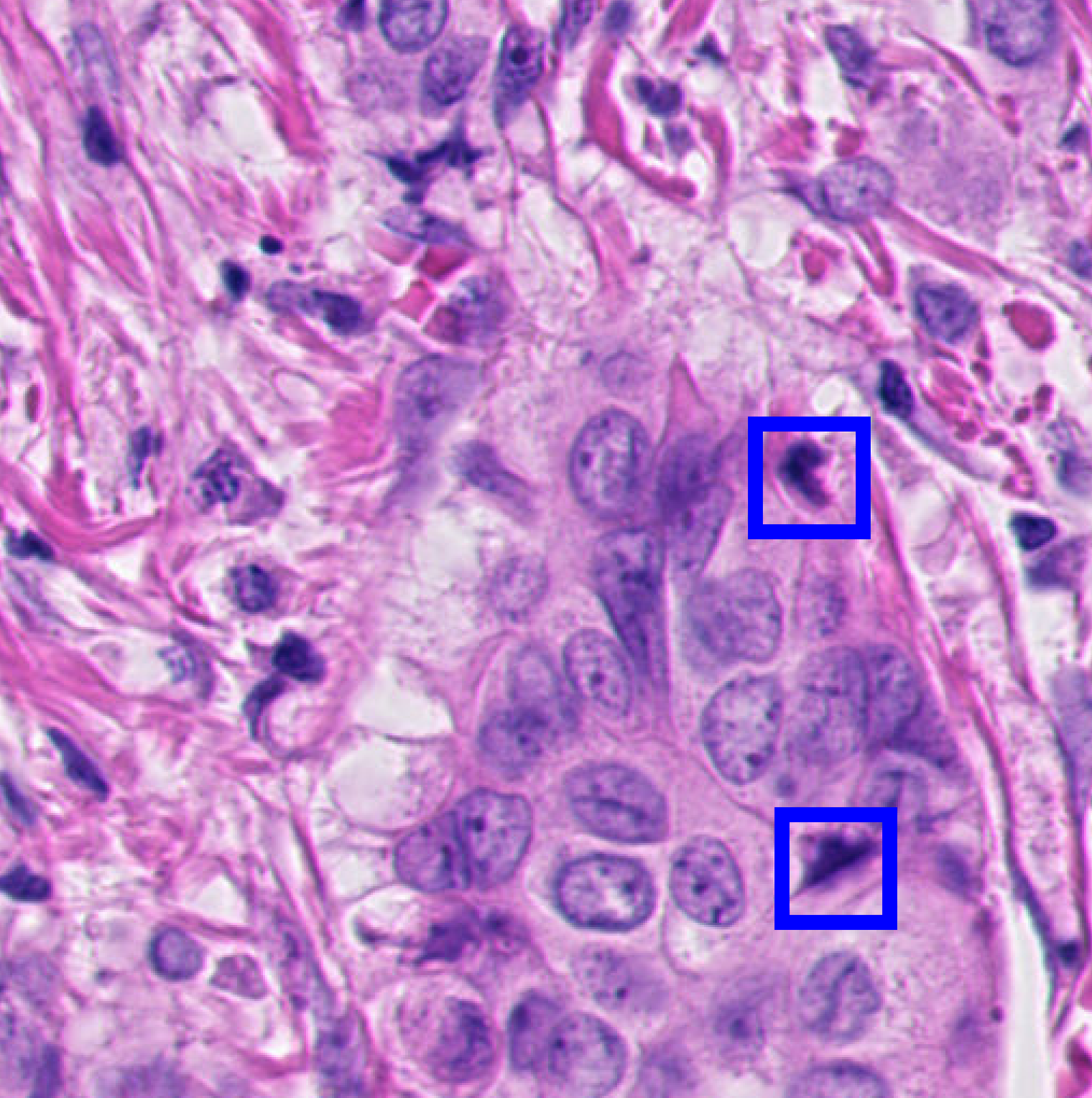}
    \caption{STRAP}
    \label{fig:strap}
\end{subfigure}
\begin{subfigure}{0.24\textwidth}
    \includegraphics[width=\textwidth]{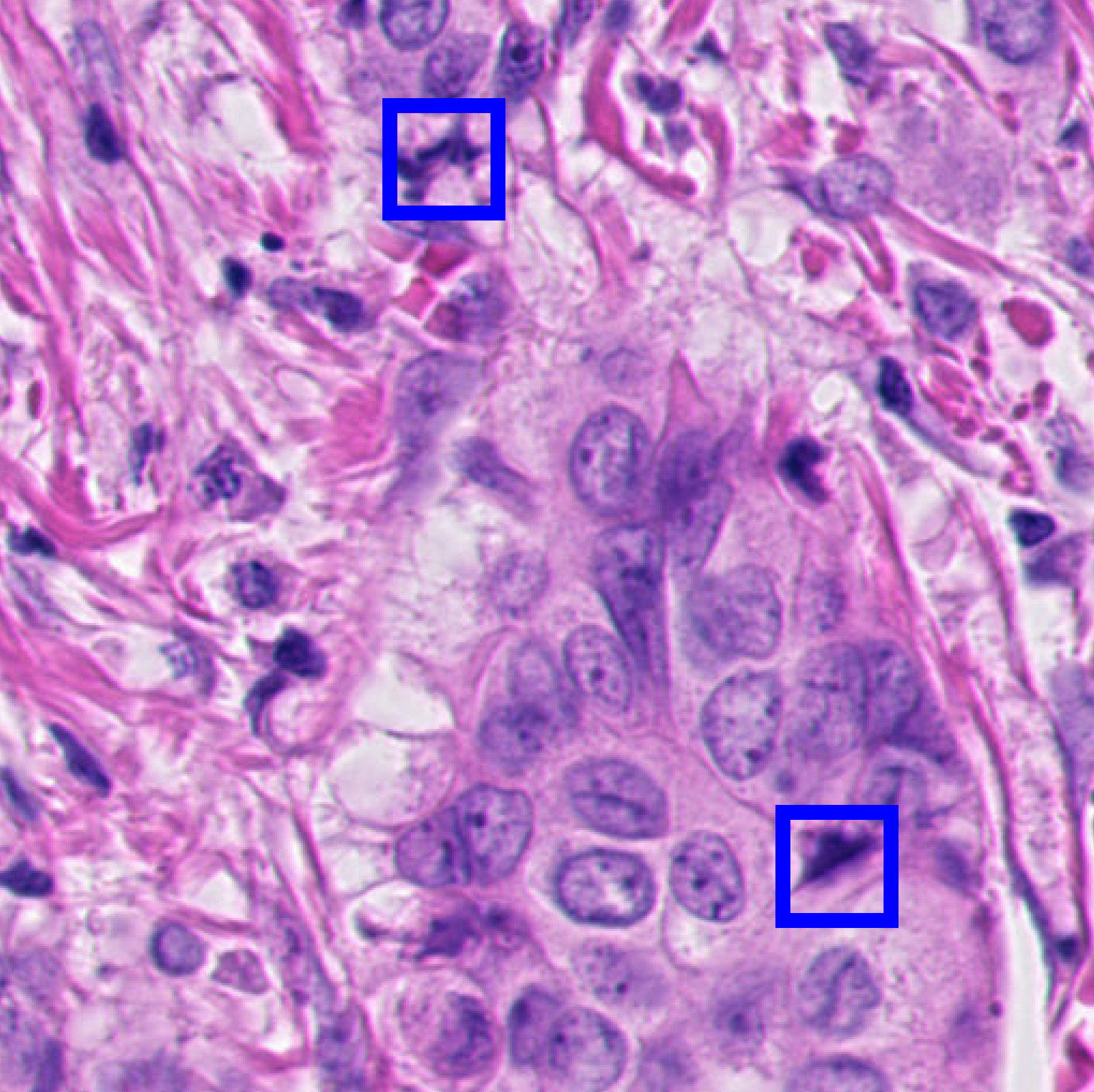}
    \caption{FuseStyle}
    \label{fig:fs}
\end{subfigure}
\label{fig:Out}
\end{figure}

\vspace{-0.75cm}
\subsection{Seen Domain: S360}
\vspace{-0.75cm}
\begin{figure}[H]
\centering
\begin{subfigure}{0.24\textwidth}
    \includegraphics[width=\textwidth]{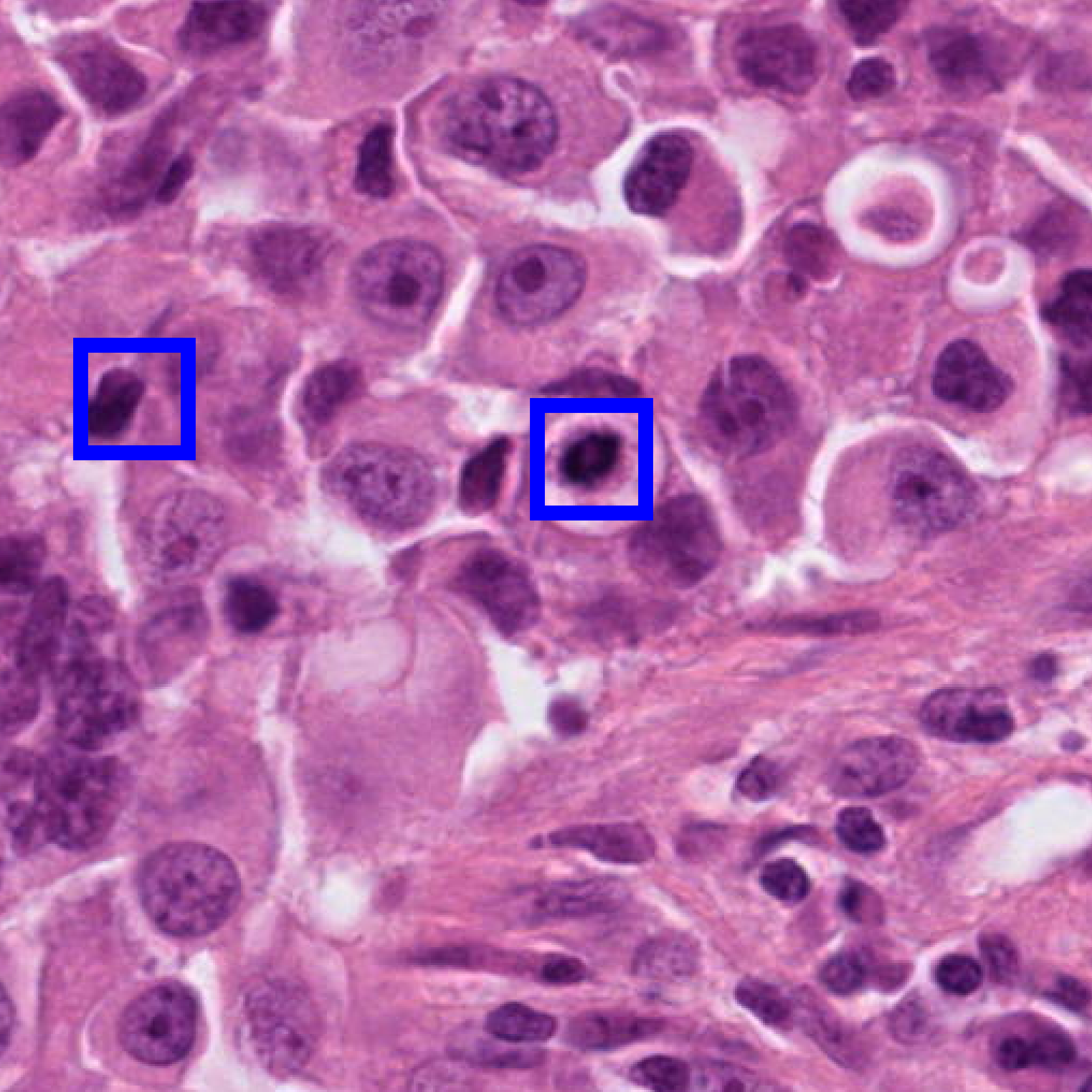}
    \caption{Ground Truth}
    \label{fig:GT}
\end{subfigure}
\begin{subfigure}{0.24\textwidth}
    \includegraphics[width=\textwidth]{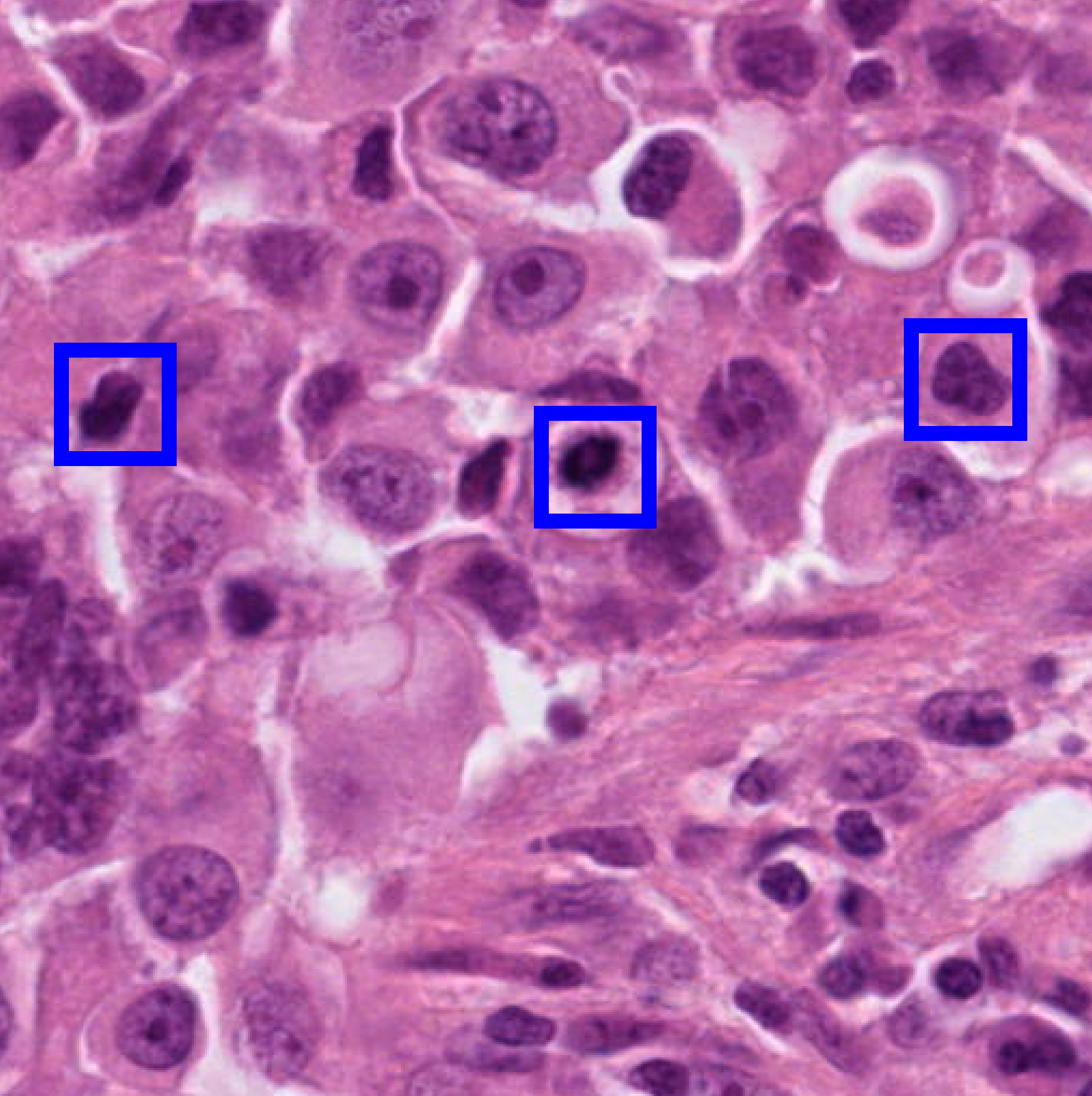}
    \caption{RetinaNet}
    \label{fig:WF}
\end{subfigure}
\begin{subfigure}{0.24\textwidth}
    \includegraphics[width=\textwidth]{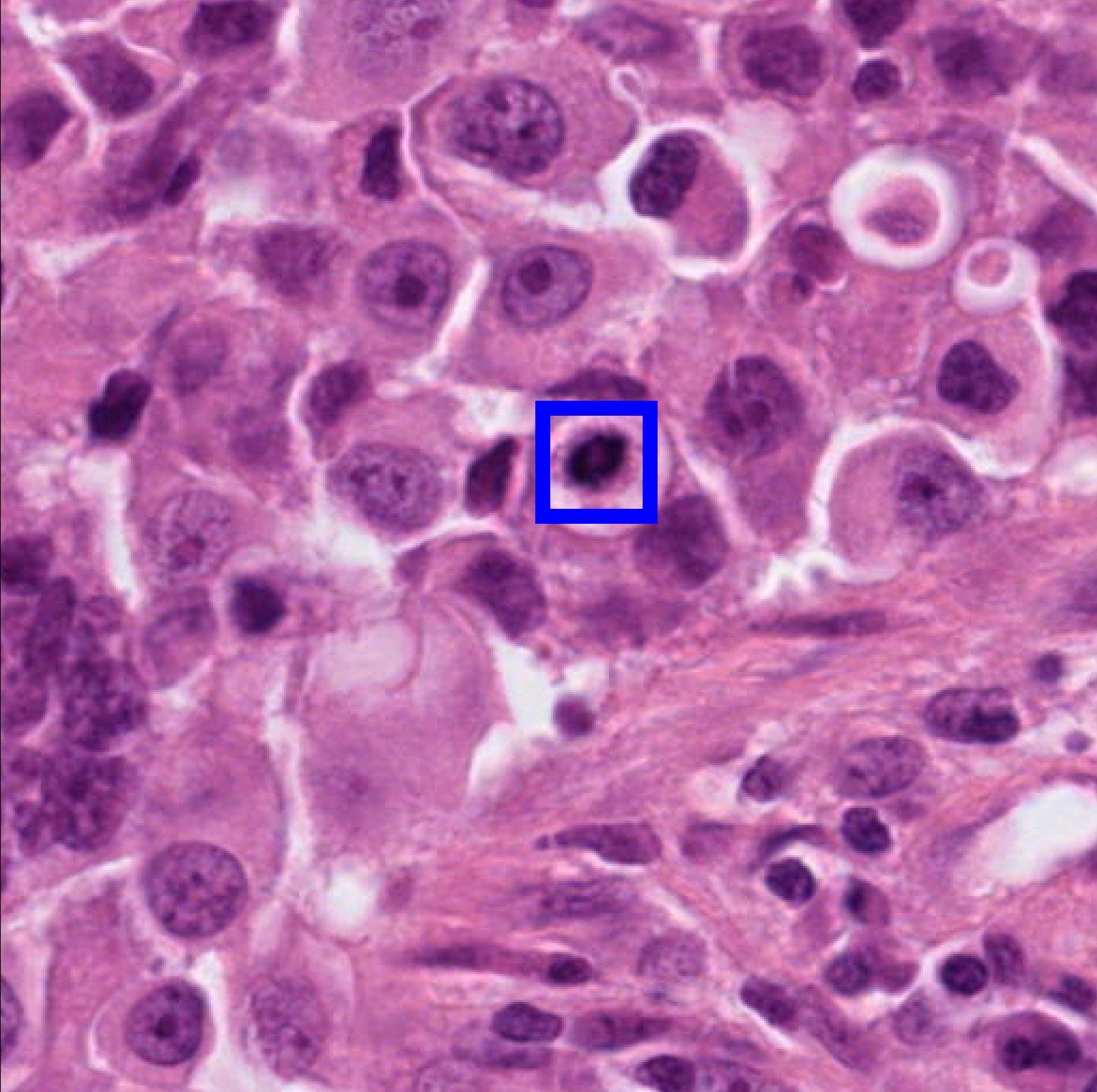}
    \caption{STRAP}
    \label{fig:strap}
\end{subfigure}
\begin{subfigure}{0.24\textwidth}
    \includegraphics[width=\textwidth]{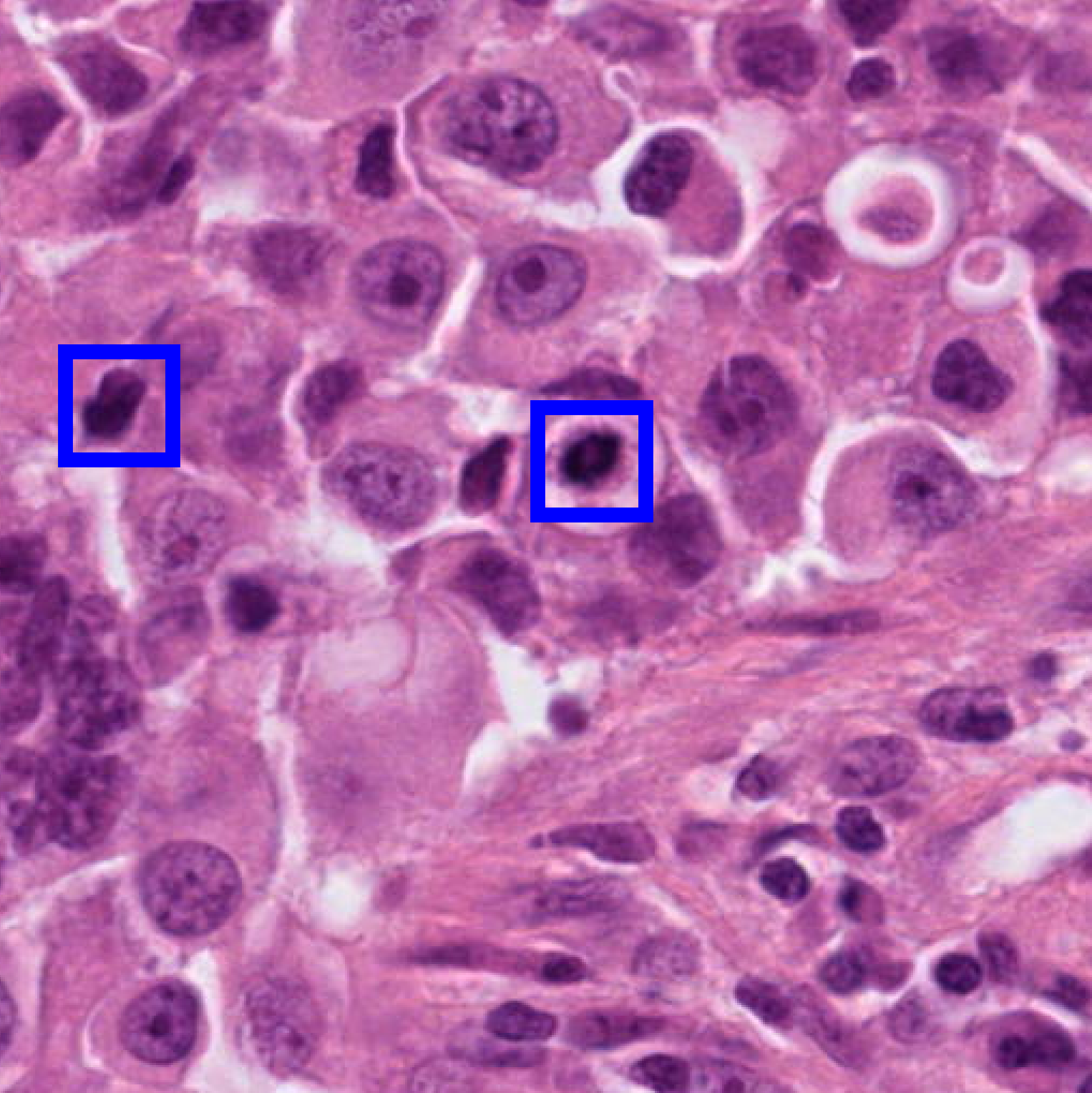}
    \caption{FuseStyle}
    \label{fig:fs}
\end{subfigure}
\label{fig:Out}
\end{figure}
\vspace{-0.5cm}
\textbf{Dataset Creation for Mitotic Figure Detection Task:}\\
\vspace{-1.5cm}
\begin{figure}[H]
    \centering
    \includegraphics[width=0.6\textwidth,height=0.4\textwidth]{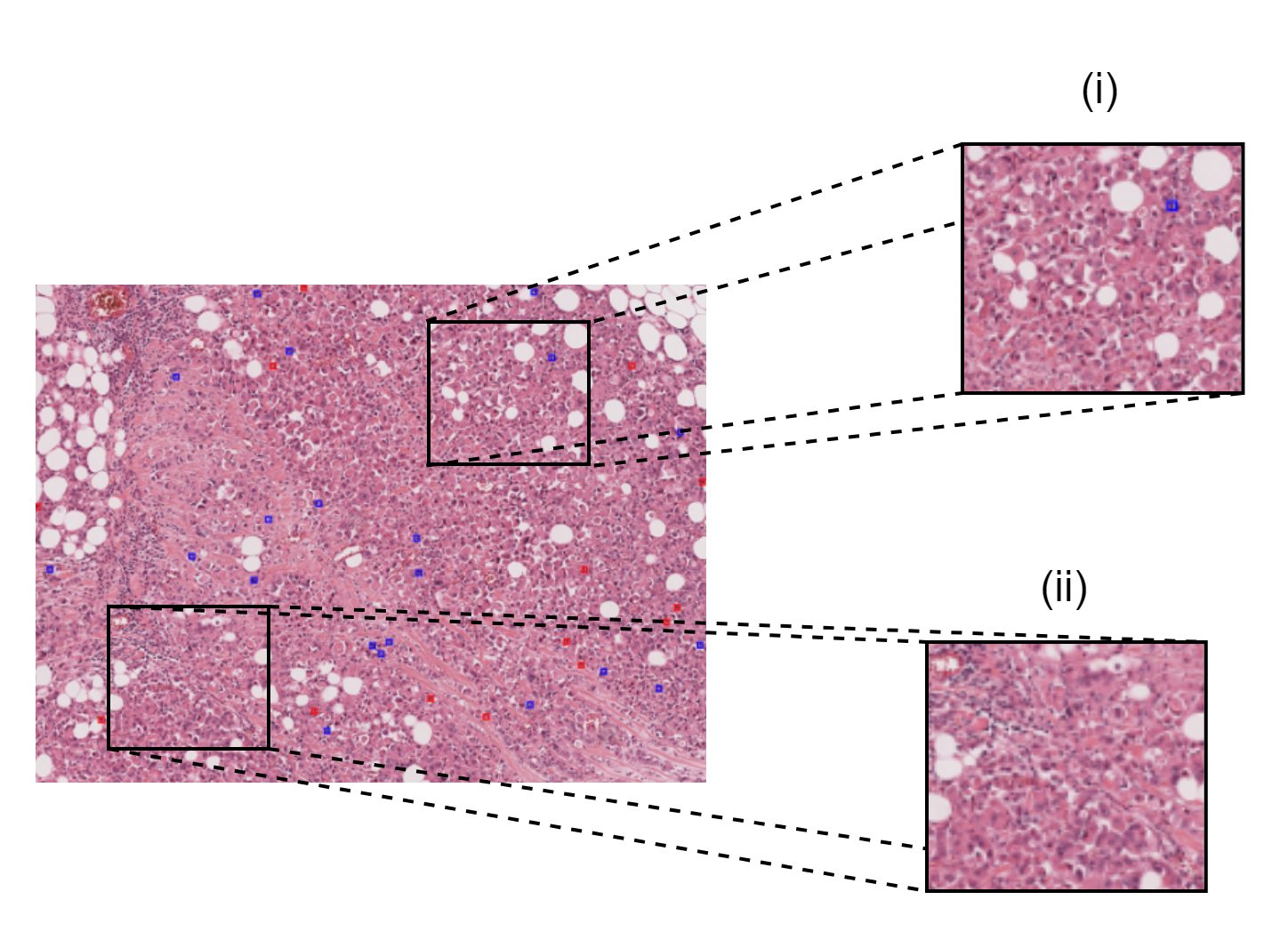}
    \caption{We extracted crops of size 512 $\times$ 512 from the whole slide images(WSI) for dataset creation for Mitotic Figure detection tasks. Following the dataset creation strategy of \cite{DAR}, we also cropped two region types from the WSI, 1. Around Mitotic Figure (Fig.\ref{fig:DS}[i]), 2. Random Crops, which may not contain any mitotic figure (Fig.\ref{fig:DS}[ii]).}
    \label{fig:DS}
\end{figure}\\
\vspace{-0.75cm}
 \bibliographystyle{splncs04}
\bibliography{ref}
%
